\documentclass{article}

\usepackage{microtype}
\usepackage{graphicx}
\usepackage{subfigure}
\usepackage{booktabs} 

\usepackage{hyperref}

\usepackage[accepted]{icml2025}

\usepackage{amsmath}
\usepackage{amssymb}
\usepackage{mathtools}
\usepackage{amsthm}

\usepackage[capitalize,noabbrev]{cleveref}

\theoremstyle{plain}
\newtheorem{theorem}{Theorem}[section]

\newtheorem{lemma}[theorem]{Lemma}

\theoremstyle{definition}

\newtheorem{assumption}[theorem]{Assumption}
\theoremstyle{remark}
\newtheorem{remark}[theorem]{Remark}

\usepackage[textsize=tiny]{todonotes}

\usepackage{comment}
\usepackage{amsmath, amsthm, amssymb}

\newcommand{\ptopl}{P2PL}
\newcommand{\hsl}{HSL}

\newcommand{\ellocal}{ELL}

\newif\ifdraft

\draftfalse

\ifdraft
\newcommand{\chop}[1]{}

\newcommand{\saurabh}[1]{\todo[inline,color=blue!40]{Saurabh: #1}}
\newcommand{\somali}[1]{\todo[inline,color=blue!40]{Somali: #1}}
\newcommand{\chaoyue}[1]{\todo[inline,color=blue!40]{Chaoyue: #1}}
\newcommand{\atul}[1]{\todo[inline,color=green!40]{Atul: #1}}
\newcommand{\kavindu}[1]{\todo[inline,color=blue!40]{Kavindu: #1}}

\else

\newcommand{\saurabh}[1]{}
\newcommand{\somali}[1]{}
\newcommand{\chaoyue}[1]{}
\newcommand{\atul}[1]{}
\newcommand{\kavindu}[1]{}

\fi

\usepackage{tikz}
\usetikzlibrary{arrows,arrows.meta,positioning}

\usepackage{placeins}

\icmltitlerunning{HSL}

\begin{document}

\twocolumn[
\icmltitle{Hubs and Spokes Learning: Efficient and Scalable Collaborative \\
           Machine Learning}



\icmlsetsymbol{equal}{*}

\begin{icmlauthorlist}
\icmlauthor{Atul Sharma}{purdue}
\icmlauthor{Kavindu Kherath}{purdue}
\icmlauthor{Saurabh Bagchi}{purdue}
\icmlauthor{Somali Chaterji}{purdue}
\icmlauthor{Chaoyue Liu}{purdue}
\end{icmlauthorlist}

\icmlaffiliation{purdue}{Elmore Family School of Electrical and Computer Engineering, Purdue University, West Lafayette, United States of America}

\icmlcorrespondingauthor{Atul Sharma}{sharm438@purdue.edu}

\icmlkeywords{collaborative machine learning, scalable, reliable, federated learning, distributed learning, decentralized learning, peer-to-peer learning, hubs and spokes learning}

\vskip 0.3in
]




\begin{abstract}
We introduce the \textit{Hubs and Spokes Learning} (HSL) framework, a novel paradigm for collaborative machine learning that combines the strengths of Federated Learning (FL) and Decentralized Learning (P2PL).
HSL employs a two-tier communication structure that avoids the single point of failure inherent in FL and outperforms the state-of-the-art P2PL framework, Epidemic Learning Local (ELL).
At equal communication budgets (total edges), HSL achieves higher performance than ELL, while at significantly lower communication budgets, it can match ELL’s performance.
For instance, with only 400 edges, HSL reaches the same test accuracy that ELL achieves with 1000 edges for 100 peers (spokes) on CIFAR-10, demonstrating its suitability for resource-constrained systems.
HSL also achieves stronger consensus among nodes after mixing, resulting in improved performance with fewer training rounds.
We substantiate these claims through rigorous theoretical analyses and extensive experimental results, showcasing HSL’s practicality for large-scale collaborative learning.
\end{abstract}
\section{Introduction}
\label{sec:intro}
\atul{Ready for review}

\begin{figure*}[tb]
\centering
\begin{tikzpicture}[
    >=stealth',
    spoke/.style={
      circle,
      draw=black,
      fill=teal!20,
      minimum size=12pt
    },
    hub/.style={
      rectangle,
      draw=black,
      fill=orange!20,
      minimum width=20pt,
      minimum height=14pt
    },
    push/.style={
      ->,
      thick,
      dashed,
      dash pattern=on 3pt off 2pt,
      color=blue
    },
    gossip/.style={
      ->,
      thick,
      dashed,
      dash pattern=on 6pt off 2pt,
      color=red
    },
    pull/.style={
      ->,
      thick,
      dashed,
      dash pattern=on 1pt off 2pt on 1pt off 2pt,
      color=purple
    },
    font=\footnotesize
]

\node[hub] (H1) at (0,3) {$H_1$};
\node[hub] (H2) at (3,2.3) {$H_2$};
\node[hub] (H3) at (6,2.3) {$H_3$};
\node[hub] (H4) at (9,3) {$H_4$};

\node[spoke] (S1) at (0,0) {$S_1$};
\node[spoke] (S2) at (1,0) {$S_2$};
\node[spoke] (S3) at (2,0) {$S_3$};
\node[spoke] (S4) at (3,0) {$S_4$};
\node[spoke] (S5) at (4,0) {$S_5$};
\node[spoke] (S6) at (5,0) {$S_6$};
\node[spoke] (S7) at (6,0) {$S_7$};
\node[spoke] (S8) at (7,0) {$S_8$};
\node[spoke] (S9) at (8,0) {$S_9$};

\draw[push] (S1) -- (H1);
\draw[push] (S4) -- (H1);
\draw[push] (S7) -- (H1);

\draw[push] (S2) -- (H2);
\draw[push] (S5) -- (H2);
\draw[push] (S8) -- (H2);

\draw[push] (S3) -- (H3);
\draw[push] (S6) -- (H3);
\draw[push] (S9) -- (H3);

\draw[push] (S1) -- (H4);
\draw[push] (S3) -- (H4);
\draw[push] (S8) -- (H4);

\draw[gossip] (H1) -- (H3);
\draw[gossip] (H2) -- (H4);
\draw[gossip] (H3) -- (H2);
\draw[gossip] (H4) -- (H1);

\draw[pull] (H3) -- (S1);
\draw[pull] (H1) -- (S2);
\draw[pull] (H4) -- (S3);
\draw[pull] (H1) -- (S4);
\draw[pull] (H4) -- (S5);
\draw[pull] (H2) -- (S6);
\draw[pull] (H2) -- (S7);
\draw[pull] (H3) -- (S8);
\draw[pull] (H1) -- (S9);

\node at (-2.5,1.5) [draw=black, fill=white, font=\scriptsize, inner sep=2pt] (legend) {
   \begin{tikzpicture}[x=0.8cm, y=0.5cm, >=stealth']
     \draw[push] (0,2)--(0.7,2);
     \node[right] at (0.7,2) {Push (S\(\to\)H)};

     \draw[gossip] (0,1)--(0.7,1);
     \node[right] at (0.7,1) {Gossip (H\(\to\)H)};

     \draw[pull] (0,0)--(0.7,0);
     \node[right] at (0.7,0) {Pull (H\(\to\)S)};
   \end{tikzpicture}
};

\end{tikzpicture}
\caption{A snapshot of the Hubs and Spokes Learning (HSL) network with 9 spokes and 4 hubs with 25 directed edges, where connections dynamically change in each round, illustrating the three-stage communication process. In \textbf{Stage 1 (Spoke-to-Hub Push)}, hubs aggregate models from \( b_{hs} = 3 \) randomly sampled spokes (depicted by blue dashed lines). In \textbf{Stage 2 (Hub Gossip)}, each hub exchanges models with \( b_{hh} = 1 \) other hub (shown in red dashed lines) and averages the received models along with its own. Finally, in \textbf{Stage 3 (Hub-to-Spoke Pull)}, each spoke retrieves a model from \( b_{sh} = 1 \) randomly selected hub (shown by purple dotted lines). 
}
\label{fig:hsl}
\end{figure*}
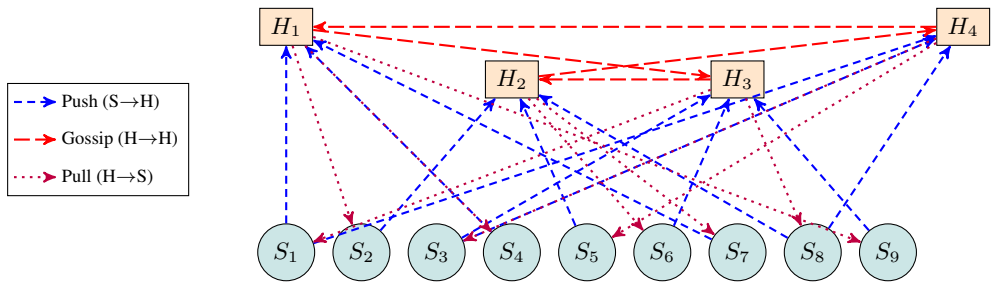

In modern machine learning, particularly in settings with edge devices, sensor networks, and large organizations, training models across distributed nodes presents significant challenges. These settings pose challenges like resource constraints, unreliable networks, and heterogeneous (non-iid) data distributions across nodes~\cite{smart_healthcare, edge_devices}. 
Federated Learning (FL) addresses these by employing a central server to aggregate model updates from clients who train locally \cite{mcmahan2017communication, pmlr-v119-karimireddy20a, ye2023feddisco}. However, FL suffers from scalability bottlenecks and a single point of failure due to its reliance on the central server. Peer-to-Peer Learning (P2PL), or decentralized learning, offers an alternative by eliminating the central server and enabling nodes to directly exchange and aggregate model updates with their neighbors \cite{lian2017can, koloskova2020unified, kong2021consensus}.

Decentralized learning benefits from increasing the connectivity of nodes, often quantified by the parameter $k$, which represents the number of neighbors each node communicates with in a given round of training. Increasing $k$ has been shown to improve convergence properties, as mathematically analyzed in the dynamic random graph-based framework, \emph{Epidemic Learning}~\cite{epidemic}, which is the state-of-the-art \ptopl\ network. However, this improvement comes at the cost of higher total communication and computation, both of which are directly proportional to the total number of edges in the system. In a \ptopl\ network with $n$ nodes and degree $k$, the total number of edges scales linearly with $n$ and with $k$. For large-scale networks, this renders fully decentralized approaches increasingly resource-intensive.

\paragraph{Hubs and Spokes Learning (HSL):}
We propose the \textit{HSL framework} integrating the \textit{hierarchical structure of FL} and the \textit{decentralized nature of P2PL}. 
FL embodies high connectivity, while P2PL eliminates the single point of failure. Leveraging the strengths of both results in a scalable and resilient collaborative learning framework,
HSL, illustrated in Figure~\ref{fig:hsl}, arranges a network into \emph{client-like spokes} and \emph{server-like hubs}. Spokes---nodes that hold private data and perform local training---communicate exclusively with hubs, while hubs form a peer-to-peer subnetwork, facilitating decentralized aggregation through gossiping. 
The spokes communicate exclusively with the hubs and not with other spokes forming a directed graph. This hierarchical design with multiple hubs as peers at the top helps mitigate FL's bottleneck while reducing the overall communication and computation costs in the system.
A key limitation of fully decentralized methods is that maintaining strong model mixing at scale requires increasing node connectivity ($k$), which inflates the communication costs. In contrast, the three subnetworks in HSL—spokes to hubs, hubs-to-hubs, and hubs-to-spokes—increase graph connectivity without requiring spokes to maintain extensive connections. By leveraging a smaller number of hubs, HSL scales efficiently as more spokes join, preventing overload. 
Moreover, independent tuning of mixing levels at the hub and spoke layers offers flexibility in model propagation, improving convergence and robustness.
Thus, HSL bridges the gap between FL and fully decentralized methods by achieving efficient model mixing without increasing the communication burden on individual nodes.
We make the following contributions:
\begin{enumerate}
    \item \textbf{Framework Definition:} We formalize the HSL design, which combines the hierarchical structure of FL with the decentralized communication of P2PL. By assigning distinct roles to hubs and spokes and enabling hubs to form a P2P subnetwork, HSL achieves efficient collaboration while mitigating FL’s bottleneck and the high cost of full decentralization.
    \item \textbf{Theoretical Analysis:} We provide rigorous convergence guarantees for HSL, demonstrating that its two-tiered structure facilitates efficient information propagation and achieves asymptotic convergence under the standard assumptions of smoothness, bounded stochastic noise, and bounded heterogeneity. 
    \item \textbf{Consensus Distance Bounds:} We 
    derive analytical bounds on the consensus distance ratio, quantifying the effectiveness of model mixing at different stages of HSL. 
    Our framework enables independent tuning of hub and spoke budgets 
    leading to improved mixing efficiency while keeping per-spoke communication costs low. By controlling the \emph{consensus distance ratio}, a measure of mixing effectiveness, HSL achieves efficient model propagation without necessarily increasing spokes' communication budgets---offering a key advantage over fully decentralized P2PL frameworks.
    
    \item \textbf{Empirical Validation:} 
    HSL achieves high accuracy even with constrained communication budgets, consistently outperforming or matching \emph{EL Local}, the SOTA P2PL framework, while using fewer edges. 

    On both the CIFAR-10 and AG News datasets, HSL with just 400 edges achieves the same local test accuracy as ELL with 1000 edges for 100 spokes. Similarly, for 200 spokes, HSL requires fewer than 600 edges to match ELL's performance with 3000 edges.
\end{enumerate}

\section{Background and Related Work}
\label{sec:related}
\atul{ready for reveiw. Kavindu, fill in the citations}
Consider a distributed learning setup with $n_s$ nodes, each holding a private dataset $\mathcal{D}_i$. The nodes collaborate to train their models while maintaining privacy by exchanging model updates rather than raw data. 
Every node initializes the same model $\mathbf{x}_0$ and follows a common training procedure.
Let \( f^{(i)}(\mathbf{x}) \) denote the local objective function optimized by node \( i \) over its private dataset \( \mathcal{D}_i \), given by:
\[
f^{(i)}(\mathbf{x}) := \mathbb{E}_{\xi \sim \mathcal{D}_i} [ f(\mathbf{x}, \xi) ],
\]
where \( \xi \) is a randomly sampled mini-batch, and the expectation is taken over the data distribution \( \mathcal{D}_i \) of node \( i \), and \( f(\mathbf{x}, \xi) \). 
The global objective is formulated as the minimization of the average local objectives across all \( n_s \) nodes:
\[
\min_{\mathbf{x}} F(\mathbf{x}) = \min_{\mathbf{x}} \frac{1}{n_s} \sum_{i=1}^{n_s} f^{(i)}(\mathbf{x}),
\]
where \( F(\mathbf{x}) \) represents the global loss function to be minimized.
We define the model state at iteration~$t$ using the matrix 
$X_t$, whose $i$-th row is $\mathbf{x}_t^{(i)}$:
\[
X_t \;:=\; 
\bigl[\mathbf{x}_t^{(1)},\;\mathbf{x}_t^{(2)},\;\dots,\;\mathbf{x}_t^{(n_s)}\bigr]^\top 
\;\in\;\mathbb{R}^{n_s\times d},
\]
where $d$ is the number of model parameters. Each node performs local training on mini-batches of $\mathcal{D}_i$, updating its model from $\mathbf{x}_t^{(i)}$ to an intermediate state $\mathbf{x}_{t'}^{(i)}$, where $t < t' < t+1$. After local training, nodes exchange (or \emph{mix}) their updated models following:
\begin{equation}
\label{eq:mixing_equation}
    X_{t+1} \;=\; W_t\, X_{t'},
\end{equation}
where $W_t \in \mathbb{R}^{n_s \times n_s}$ is the \emph{mixing matrix} at iteration~$t$.
The mixing matrix $W_t$ governs the information exchange, determining how nodes aggregate updates from their neighbors.

\paragraph{Federated Learning (FL):} 
One approach to distributed training with decentralized data is \emph{federated averaging}~\cite{mcmahan2017communication}, where a central server 
computes a weighted average of client updates, weighted by local data sizes, and broadcasts the global model. 
In the notation of~\eqref{eq:mixing_equation}, the mixing matrix $W_t$ in FL round has identical rows, ensuring all clients receive the same update.
While this star topology enables exact consensus with only $n_s$ edges, it creates a single point of failure and burdens the central server as clients scale. A sampling scheme can reduce this load but slows training.

\paragraph{Peer-to-Peer Learning (\ptopl):}
In P2PL, the nodes handle mixing themselves, optimizing the same global objective via Decentralized-SGD (D-SGD). The mixing matrix $W_t$ encodes adjacency relationships, dictating how nodes ``gossip''. Unlike FL, which enforces exact consensus via a server, P2PL follows \emph{inexact consensus}, meaning models $\mathbf{x}^{(i)}$ are not necessarily identical at any given time. 
The \emph{consensus distance} (CD) measures the average Euclidean distance of individual models from their mean: 
\begin{equation}
CD_t = \frac{1}{n_s} \sum_{i=1}^{n_s} \left\|\mathbf{x}_t^{(i)} - \bar{\mathbf{x}}_t\right\|^2,
\end{equation}
where $\bar{\mathbf{x}}_t = \frac{1}{n_s} \sum_{i=1}^{n_s} \mathbf{x}_t^{(i)}$ is the mean model at round $t$. 
Mixing reduces $CD_t$, aligning models across nodes. The \emph{CD ratio} (CDR) quantifies mixing effectiveness: 
\begin{equation}
\text{CDR} = \frac{CD_{t+1}}{CD_{t'}}.
\end{equation}
Lower CDR indicates stronger mixing and improved model alignment. 
Prior works~\cite{lian2017can} considered static mixing matrices, while~\cite{assran2019stochastic, koloskova2020unified, kong2021consensus} perform D-SGD for dynamic graphs, often assuming a doubly stochastic mixing matrix, which is impractical in full decentralization. 
Recent work has shown decentralized optimization with non-doubly stochastic matrices is viable, broadening mixing protocols.
\emph{Epidemic Learning (EL)}~\cite{epidemic}, the SOTA approach, improves convergence by dynamically changing communication topologies. Each node sends updates to a random peer subset, outperforming static or structured methods in both speed and accuracy.

\paragraph{Epidemic Learning (EL):}
EL has two variants: \emph{EL Oracle} and \emph{EL Local}. EL Oracle enforces a $k$-regular dynamic graph, where every node maintains both in-degree and out-degree of exactly $k$, requiring a central coordinator, which conflicts with full decentralization. To address this, EL Local is designed to ensure decentralization, where each node maintains a fixed out-degree of $k$, sharing its model updates with $k$ peers, while its in-degree can vary around the expected value $k$. Despite this relaxation, EL Local retains EL's strong mixing and convergence properties, enabling decentralized learning without central coordination.
\atul{Todo-mention directed vs undirected graphs}

\paragraph{Other Approaches:} 
Other works have explored architectures like hierarchical FL~\cite{liu2020client, abad2020hierarchical}, which introduces intermediary aggregators but retains a single point of failure at the root server. Blockchain-based decentralized learning~\cite{korkmaz2020chain, qin2024blockdfl} enhances consensus but incurs significant overhead. Additionally, works such as~\cite{dhasade2023decentralized, beltran2023decentralized} address scalability challenges in collaborative machine learning networks, including communication bottlenecks and system efficiency as networks grow.

\section{Hubs and Spokes Learning (\hsl)}
\label{sec:design}
\atul{Ready for review}
Now that we have established how FL and P2PL enable collaboration among nodes, we introduce our \emph{hubs and spokes learning} (\hsl) framework, which facilitates collaboration through intermediate \emph{hubs}.
In \hsl, \emph{spokes} are client-like nodes that perform local training, while \emph{hubs} serve as intermediaries that aggregate and mix updates. Each spoke communicates exclusively with hubs—sending 
model updates and receiving aggregated models---without direct spoke-to-spoke communication. 
This structure, illustrated in Figure~\ref{fig:hsl}, 
improves efficiency by offloading mixing to hubs while decentralizing them ensures scalability and removes FL's single point of failure.

\paragraph{Training in \hsl:}  
At each round \(t\), spoke \(i\) begins with its model \(x_t^{(i)}\) and performs a few local SGD rounds on its private loss function \(F_i(\cdot)\), updating to \(x_{t+\tfrac{1}{4}}^{(i)}\). The collection of spoke updates is:  
\begin{align*}
X_{t+\tfrac{1}{4}}
=
\left[
x_{t+\tfrac{1}{4}}^{(1)},\, 
x_{t+\tfrac{1}{4}}^{(2)},\,\dots,\,
x_{t+\tfrac{1}{4}}^{(n_s)}
\right]^\top
\in
\mathbb{R}^{n_s \times d}.
\end{align*}

Each communication round proceeds in three steps:
\begin{enumerate}
    \item \textbf{Spoke-to-Hub Push:} 
    Each hub \(k\) requests the updated models from \(b_{hs}\) spokes, forming the spoke-to-hub mixing matrix \(W_{hs} \in \mathbb{R}^{n_h \times n_s}\). 
    Concretely, let \(\mathcal{S}_k\) be the set of \(b_{hs}\) spokes from which hub \(k\) collects models. 
    Hub \(k\) then aggregates these models as:
    \[
    x_{t+\tfrac{2}{4}}^{(k)}
    =
    \frac{1}{b_{hs}}
    \sum_{i \in \mathcal{S}_k}
    x_{t+\tfrac{1}{4}}^{(i)},
    \quad
    X_{t+\tfrac{2}{4}}
    = 
    W_{hs}\,X_{t+\tfrac{1}{4}},
    \]
    where \(X_{t+\tfrac{2}{4}} \in \mathbb{R}^{n_h\times d}\) represents all hub models after this step. 
    Here, each hub has a fixed in-degree of \(b_{hs}\), as it collects models from the same number of spokes. The out-degree of spokes, however, may vary around an expected value of \((n_h \cdot b_{hs}) / n_s\), as spokes share their models with multiple hubs.
    \item \textbf{Hub-to-Hub Gossip:} 
    The hubs form a peer-to-peer network and execute the gossip scheme from \emph{EL Local}, where each hub shares its model with \(b_{hh}\) other hubs. \kavindu{Should we mention EL Local here? or just say the method} 
    \atul {I think it is okay to mention}
    While the out-degree is fixed at \(b_{hh}\), the in-degree varies around an average of \(b_{hh}\) since it depends on which hubs receive models. Let \(\mathcal{A}_k\) denote the set of hubs from which \(k\) receives models. The updated hub model is given by:
    \[
    x_{t+\tfrac{3}{4}}^{(k)}
    =
    \frac{1}{|\mathcal{A}_k| + 1}
    \left(
    x_{t+\tfrac{2}{4}}^{(k)}
    +
    \sum_{m \in \mathcal{A}_k}
    x_{t+\tfrac{2}{4}}^{(m)}
    \right),
    \]
    \[
    X_{t+\tfrac{3}{4}}
    = 
    W_{hh}\,X_{t+\tfrac{2}{4}},
    \]

    where \(W_{hh} \in \mathbb{R}^{n_h \times n_h}\) is the hub-hub mixing matrix.
    \item \textbf{Hub-to-Spoke Pull:}
    In the final step, each spoke \(i\) queries a random set \(\mathcal{H}_i\) of \(b_{sh}\) hubs and averages their models:
    \[
    x_{t+1}^{(i)}
    =
    \frac{1}{b_{sh}}
    \sum_{k \in \mathcal{H}_i}
    x_{t+\tfrac{3}{4}}^{(k)},
    \quad
    X_{t+1}
    =
    W_{sh}\,X_{t+\tfrac{3}{4}},
    \]
    where \(W_{sh} \in \mathbb{R}^{n_s\times n_h}\) is the hub-to-spoke mixing matrix. 
    Each spoke has a fixed in-degree \(b_{sh}\), while hubs may have a variable out-degree in this step, with an expected value of \((n_s \cdot b_{sh}) / n_h\).
\end{enumerate}

We summarize the training process in Algorithm~\ref{alg:hsl}.
From the spokes’ perspective, the three aggregation steps yield an \emph{end-to-end} transformation:
\begin{equation}
X_{t+1}
=
\left(W_{sh}\,W_{hh}\,W_{hs}\right)\,
X_{t+\tfrac{1}{4}}
\;\equiv\;
W_{hsl}\,X_{t+\tfrac{1}{4}},
\end{equation}
where \(W_{hsl} (\in \mathbb{R}^{n_s \times n_s}) = W_{sh}\,W_{hh}\,W_{hs}\) is the overall mixing matrix for one round of \hsl.
Although individual mixing matrices $W_{sh}$, $W_{hh}$, and $W_{hs}$ may be sparse due to communication budget constraints (i.e., $b_{hs}, b_{hh}, b_{sh}$), the effective matrix $W_{hsl} = W_{sh}\,W_{hh}\,W_{hs}$ is typically not sparse. Empirically, \hsl\ exhibits a larger spectral gap than \ptopl\ at comparable budgets, signifying stronger connectivity and more effective information propagation between spokes.
The choice of hubs $n_h$ in \hsl\ balances individual and total communication budgets. Fewer hubs increase the load per hub, requiring each to maintain connections with more spokes. In the extreme case of unlimited individual budgets, \hsl\ reduces to FL. Conversely, increasing $n_h$ distributes load and improves fault tolerance but raises overall communication and computation costs. 
\atul{todo}

\begin{algorithm}[tb]
   \caption{Hubs and Spokes Learning (HSL)}
   \label{alg:hsl}
\begin{algorithmic}
   \STATE {\bfseries Input:} $n_s, T, l, b_{hs}, b_{hh}, b_{sh}, \eta$ 
   \STATE \textbf{Initialize:} $x_0^{(i)} \gets x_0, \quad \forall i \in [n_s]$
   \FOR{$t = 0$ to $T-1$}
       \STATE \textbf{Step 1: Local Training}  
       \STATE \quad $x_{t+\tfrac14}^{(i)} \gets$ SGD$(x_t^{(i)}, \mathcal{D}_i, l, \eta)$  
       
       \STATE \textbf{Step 2: Spoke-to-Hub Push}  
       \STATE \quad Each hub $k$ samples $\mathcal{S}_k$ with $|\mathcal{S}_k|=b_{hs}$  
       \STATE \quad $
       x_{t+\tfrac{2}{4}}^{(k)} 
       = 
       \frac{1}{b_{hs}} 
       \sum_{i \in \mathcal{S}_k} 
       x_{t+\tfrac14}^{(i)}
       $

       \STATE \textbf{Step 3: Hub-to-Hub Gossip}  
       \STATE \quad Each hub $k$ samples $\mathcal{A}_k$ with $|\mathcal{A}_k|=b_{hh}$  
       \STATE \quad $
       x_{t+\tfrac{3}{4}}^{(k)} 
       = 
       \frac{1}{|\mathcal{A}_k| + 1}
       \Bigl(
       x_{t+\tfrac{2}{4}}^{(k)} 
       + 
       \sum_{m \in \mathcal{A}_k} 
       x_{t+\tfrac{2}{4}}^{(m)}
       \Bigr)
       $

       \STATE \textbf{Step 4: Hub-to-Spoke Pull}  
       \STATE \quad Each spoke $i$ samples $\mathcal{H}_i$ with $|\mathcal{H}_i|=b_{sh}$  
       \STATE \quad $
       x_{t+1}^{(i)} 
       = 
       \frac{1}{b_{sh}}
       \sum_{k \in \mathcal{H}_i}
       x_{t+\tfrac{3}{4}}^{(k)}
       $
   \ENDFOR
   \STATE \textbf{Output:} $\{x_T^{(i)}\}_{i=1}^{n_s}$
\end{algorithmic}
\end{algorithm}

Next, we analyze its convergence properties under standard assumptions in stochastic optimization, establishing theoretical guarantees on learning performance.

\section{Theoretical Analysis}

\subsection{Convergence of HSL}

In this section, we analyze the convergence behavior of HSL under standard assumptions commonly used in stochastic first-order methods. Specifically, we assume the following:
1) \textbf{Smoothness:} capturing how the gradients change across the loss landscape,  
(2) \textbf{Stochastic Noise Bound:} controlling the randomness in gradient estimates due to mini-batch sampling, and  
(3) \textbf{Bounded Heterogeneity:} quantifying how much local objective functions deviate from the global objective. 
Formally, we state:
\chaoyue{: (1) we need to clarify the notations $f$, $f^{(i)}$, and their relation to $F$ and $F_i$. (2) when does $f$ take $\xi$ as an argument, and when does not; (3) mismatch notations $D_i$ and $D^{(i)}$.}
\atul{Noted, to be done.}
\begin{assumption}[Smoothness]\label{ass:smoothness}
For each \(i \in [n_s]\), the function \(f^{(i)}: \mathbb{R}^d \to \mathbb{R}\) is differentiable, and there exists a constant \(L < \infty\) such that for all \(x, y \in \mathbb{R}^d\):
\[
\|\nabla f^{(i)}(y) - \nabla f^{(i)}(x)\| \le L \|y - x\|.
\]
\end{assumption}

\begin{assumption}[Bounded Stochastic Noise]\label{ass:bounded_noise}
There exists a constant \(\sigma < \infty\) such that for all \(i \in [n]\) and \(x \in \mathbb{R}^d\):
\[
\mathbb{E}_{\xi \sim D^{(i)}} \!\bigl[\|\nabla f(x, \xi) - \nabla f^{(i)}(x)\|^2\bigr] \le \sigma^2.
\]
\end{assumption}

\begin{assumption}[Bounded Heterogeneity]\label{ass:bounded_heterogeneity}
There exists a constant \(\mathcal{H} < \infty\) such that for all \(x \in \mathbb{R}^d\):
\[
\frac{1}{n_s} \sum_{i \in [n_s]} \|\nabla f^{(i)}(x) - \nabla F(x)\|^2 \le \mathcal{H}^2,
\]
where \( F(\mathbf{x}) \) denotes the average of the local objective functions \( f^{(i)}(x) \). The term \( \sigma \) captures variance introduced by stochastic gradients, while \( \mathcal{H} \) quantifies the heterogeneity arising from the non-iid data distribution.

\end{assumption}

We now present our main convergence result.

\chaoyue{: Is it possible to define $\Delta_0=F(x_0) - min_{x \in \mathbb{R}^d} F(x)$, instead of $\le $}
\atul{Yes, that sounds good, will do}
\begin{theorem}
Consider Algorithm~\ref{alg:hsl} under the above assumptions. 
Let the initial optimization gap be:
$$
\Delta_0 := F(x_0) - \min_{x \in \mathbb{R}^d} F(x).
$$
Then, for any $T \geq 1$, with $n_s \geq 2$ spokes, a communication budget of $b_{sh} \geq 1$, and $n_h \geq 2$ hubs with budgets $b_{hs} \geq 1, b_{hh} \geq 1$, selecting the step size as:
\begin{align*}
\gamma \in \Theta \left( \min \left\{ 
    \sqrt{\frac{n_s \Delta_0}{TL ((1 + \beta') \sigma^2 + \beta' \mathcal{H}^2) }}
    \right. \right. \\
    \left. \left.
    \qquad \qquad \sqrt[3]{\frac{\Delta_0}{TL^2 \beta_{HSL} (\sigma^2+\mathcal{H}^2)}}, \frac{1}{L} 
\right\} \right).
\end{align*}

we have
\begin{align*}
    &\frac{1}{n_s T} \sum_{t=0}^{T-1} \sum_{i=1}^{n_s} \mathbb{E} \left[ \left\| \nabla F(x_t^{(i)}) \right\|^2 \right] \\
    &\in \mathcal{O}\bigg(\sqrt{\frac{L \Delta_0}{T n_s} ((1+\beta')\sigma^2 + \beta'\mathcal{H}^2)} \\
    &+ \sqrt[3]{\frac{L^2 \beta_{HSL} \Delta_0^2 (\sigma^2 + \mathcal{H}^2)}{T^2}} + \frac{L\Delta_0}{T}\bigg).
\end{align*}

where 
\begin{align*}
\beta_{HSL} &:= \beta_{sh} \beta_{hh} \beta_{hs} \\
\beta' &:= \frac{1}{2} \left[ \beta_{HSL} + \frac{n_s}{n_h} \beta_{hs} \left(1 + \beta_{hh}\right) \right] 
\end{align*}
and 
\begin{align*}
\beta_{hs} &:= \frac{1}{b_{hs}} \left(1 - \frac{b_{hs}-1}{n_s-1} \right) \\
\beta_{hh} &:= \frac{1}{b_{hh}} \left(1 - \left(1 - \frac{b_{hh}}{n_h-1}\right)^{n_h} \right) - \frac{1}{n_h-1} \\
\beta_{sh} &:= \frac{1}{b_{sh}} \left(1 - \frac{b_{sh}-1}{n_h-1} \right) 
\end{align*}
\end{theorem}

Each of $\beta_{hs}, \beta_{hh}, \beta_{sh}$ represents an upper bound on the CDR for the respective mixing steps in \hsl. For details, refer to~\cref{lemma: 1b} in the Appendix.

\subsection{Consensus Distance Ratio in \hsl}  

\cref{lemma: 1b} relates the expected consensus distance (CD) before and after each mixing step in \hsl. Specifically, the following inequalities hold:  
\[
\frac{\mathbb{E}[\mathrm{CD}_{t+\tfrac{2}{4}}]}{\mathbb{E}[\mathrm{CD}_{t+\tfrac{1}{4}}]} \leq \beta_{hs}, 
\frac{\mathbb{E}[\mathrm{CD}_{t+\tfrac{3}{4}}]}{\mathbb{E}[\mathrm{CD}_{t+\tfrac{2}{4}}]} \leq \beta_{hh}, 
\frac{\mathbb{E}[\mathrm{CD}_{t+1}]}{\mathbb{E}[\mathrm{CD}_{t+\tfrac{3}{4}}]} \leq \beta_{sh},
\]
which combine to yield:
$
\frac{\mathbb{E}[\mathrm{CD}_{t+1}]}{\mathbb{E}[\mathrm{CD}_{t+\frac{1}{4}}]} \leq \beta_{HSL},  
$

These \(\beta\)-values represent upper bounds on the expected consensus distance ratio (CDR) at each stage of mixing and can be tuned by adjusting the budgets \(b_{hs}, b_{hh}, b_{sh}\).

In \emph{EL Local}, the CDR satisfies \(\beta_{EL} \in \mathcal{O}(1/k)\), meaning that the only guaranteed way to improve mixing is by increasing the node budget \(k\). In contrast, hubs improve mixing without increasing the burden on individual spokes, enabling a more efficient communication-cost trade-off.

\section{Evaluation}
\label{sec:evaluation}
\atul{Ready for review}
\begin{figure*}[ht]
\centering
\begin{minipage}[t]{0.48\textwidth}
    \includegraphics[width=\textwidth]{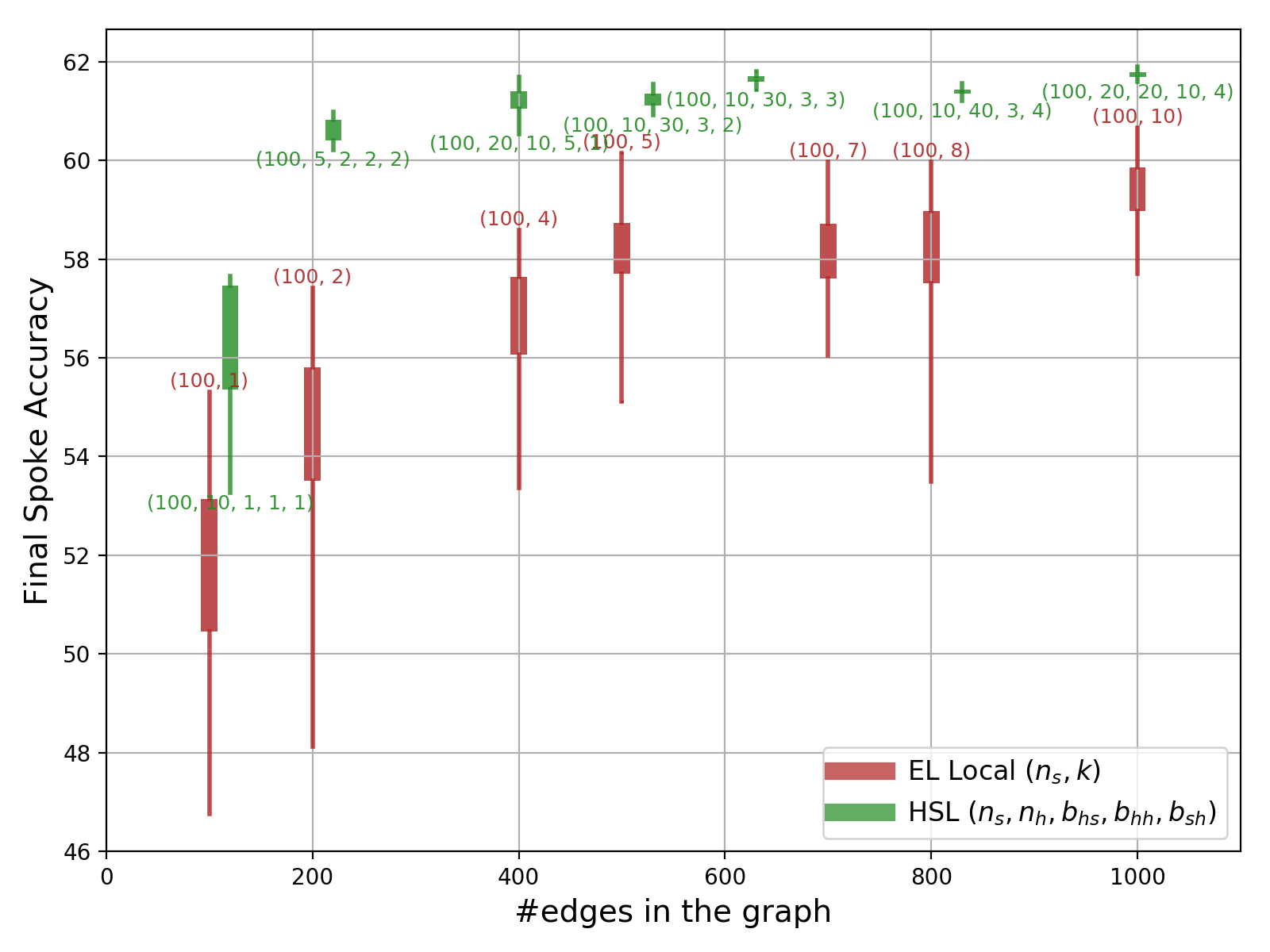}
\end{minipage}
\hfill
\begin{minipage}[t]{0.48\textwidth}
    \includegraphics[width=\textwidth]{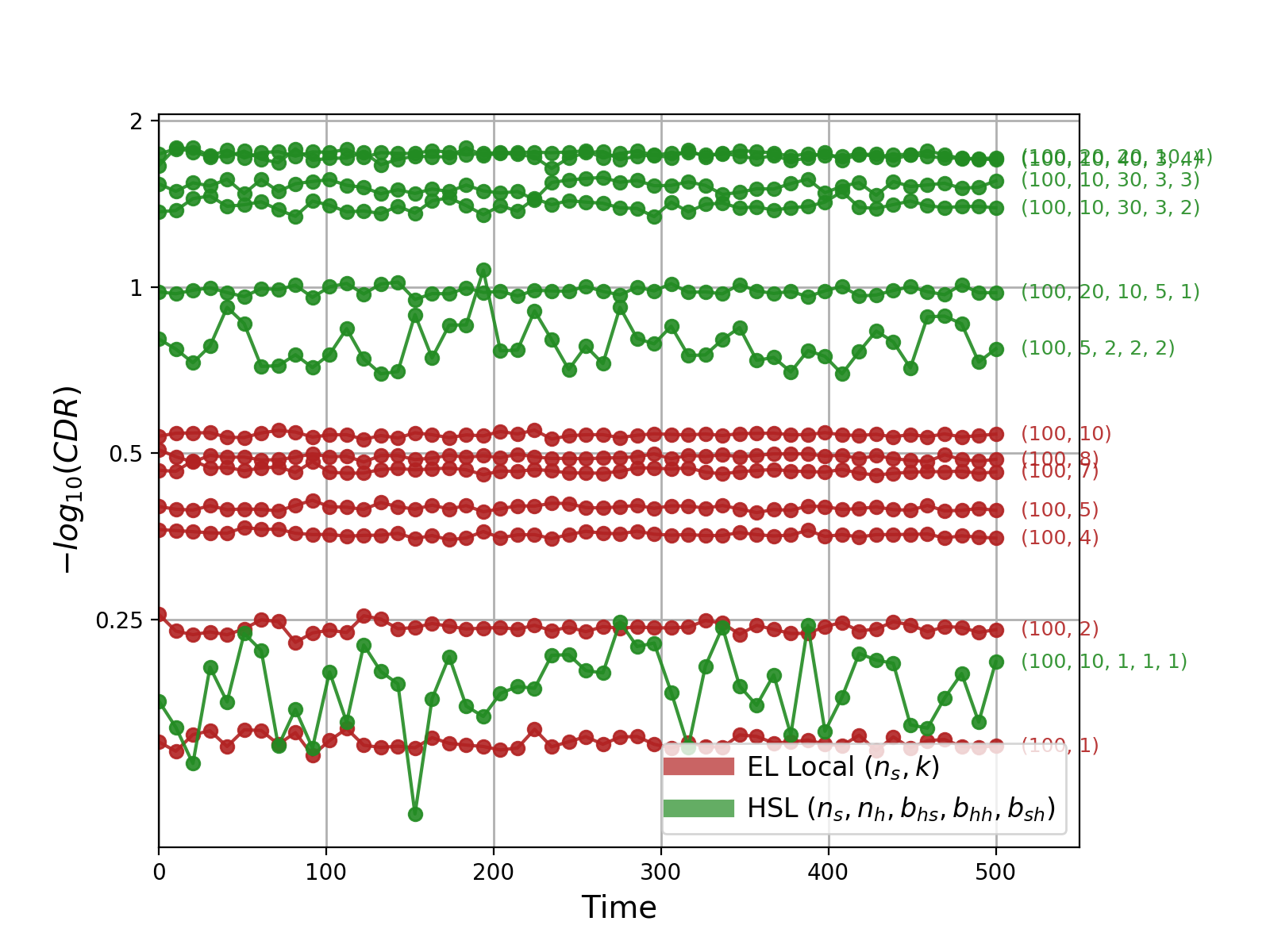}
\end{minipage}

\begin{minipage}[t]{0.48\textwidth}
    \includegraphics[width=\textwidth]{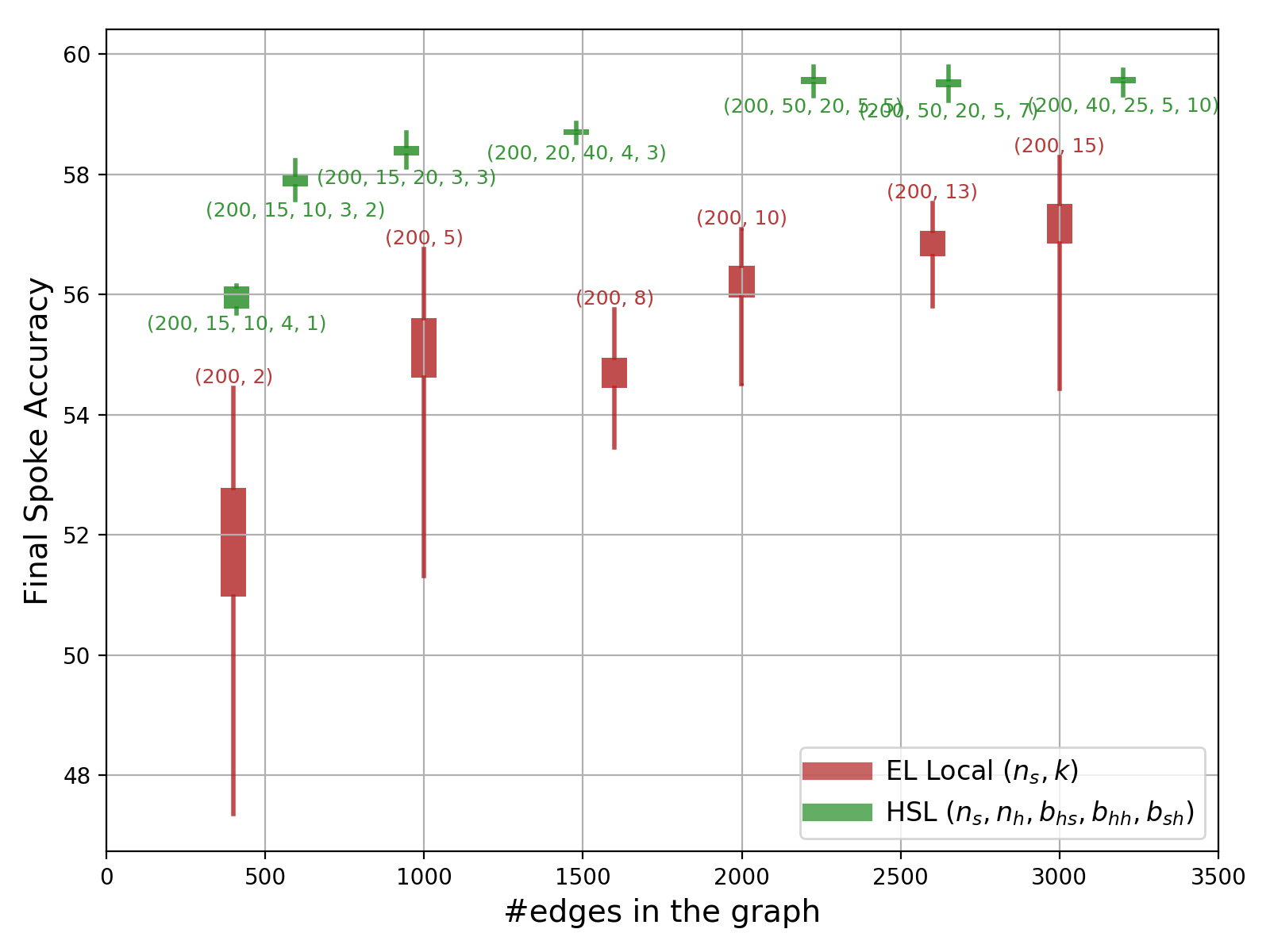}
\end{minipage}
\hfill
\begin{minipage}[t]{0.48\textwidth}
    \includegraphics[width=\textwidth]{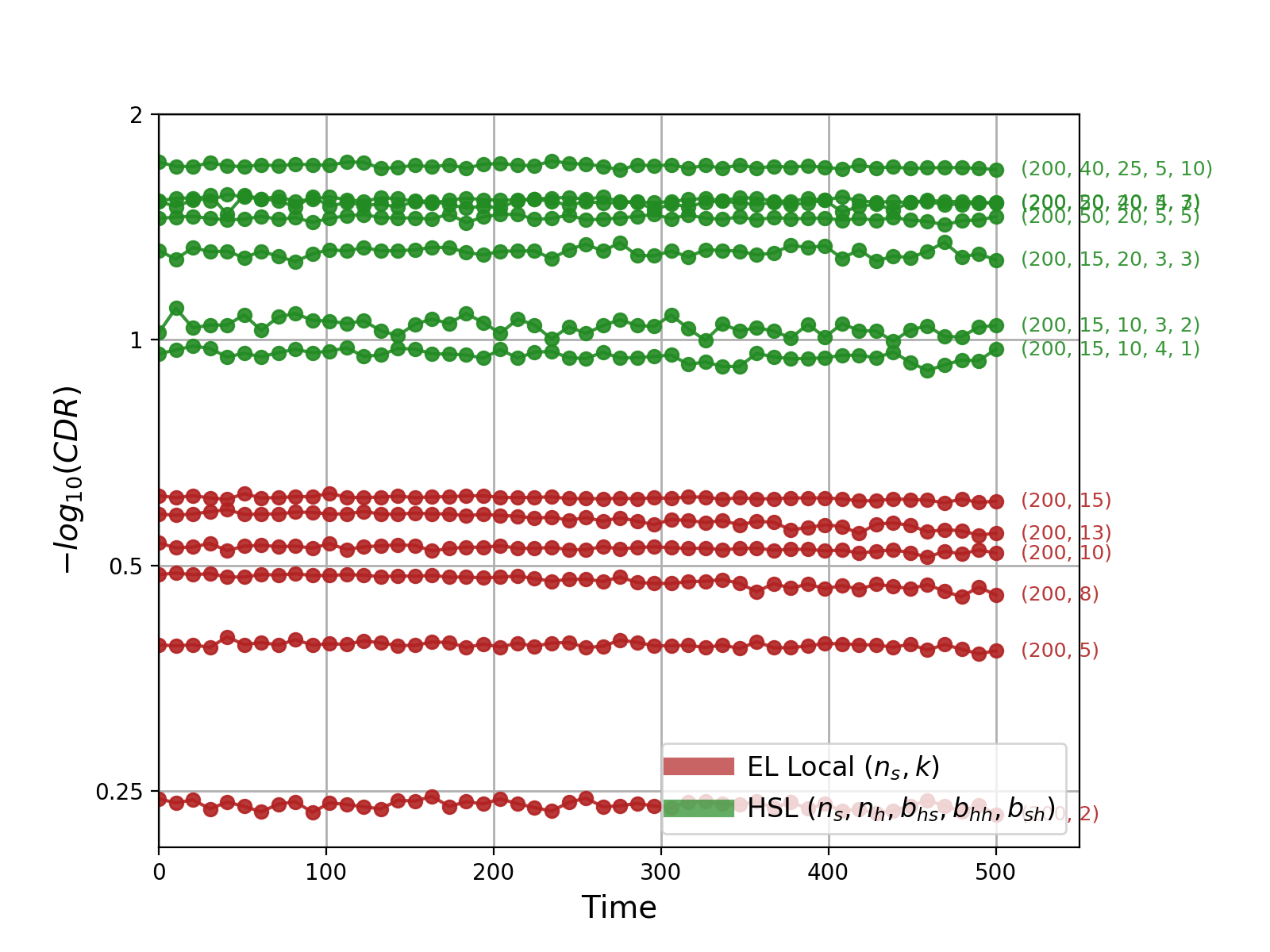}
\end{minipage}

\caption{\textbf{HSL vs.\ ELL on CIFAR-10 (\(n_s=100, 200\)).} 
(Top) \(n_s=100\): (Left) Final accuracy vs. total edges (budget). Candle bodies represent the interquartile range, and wicks indicate min/max values. HSL consistently achieves higher accuracy at lower budgets, demonstrating its efficiency and scalability. (Right) Mixing efficiency over 500 rounds, measured via \(-\log(\text{CDR})\), where higher values indicate stronger mixing. HSL achieves superior mixing at all budgets, explaining its improved accuracy.
(Bottom) \(n_s=200\): HSL maintains its advantage, matching ELL’s 3000-edge accuracy with only a third of the budget. The CDR plot further highlights HSL’s superior mixing, where HSL with just 595 edges achieves better consensus than ELL with 3000 edges, reinforcing its scalability.}

\label{fig:bud-acc-cifar}
\end{figure*}

\begin{figure*}[ht]
\centering
\includegraphics[width=0.48\textwidth]{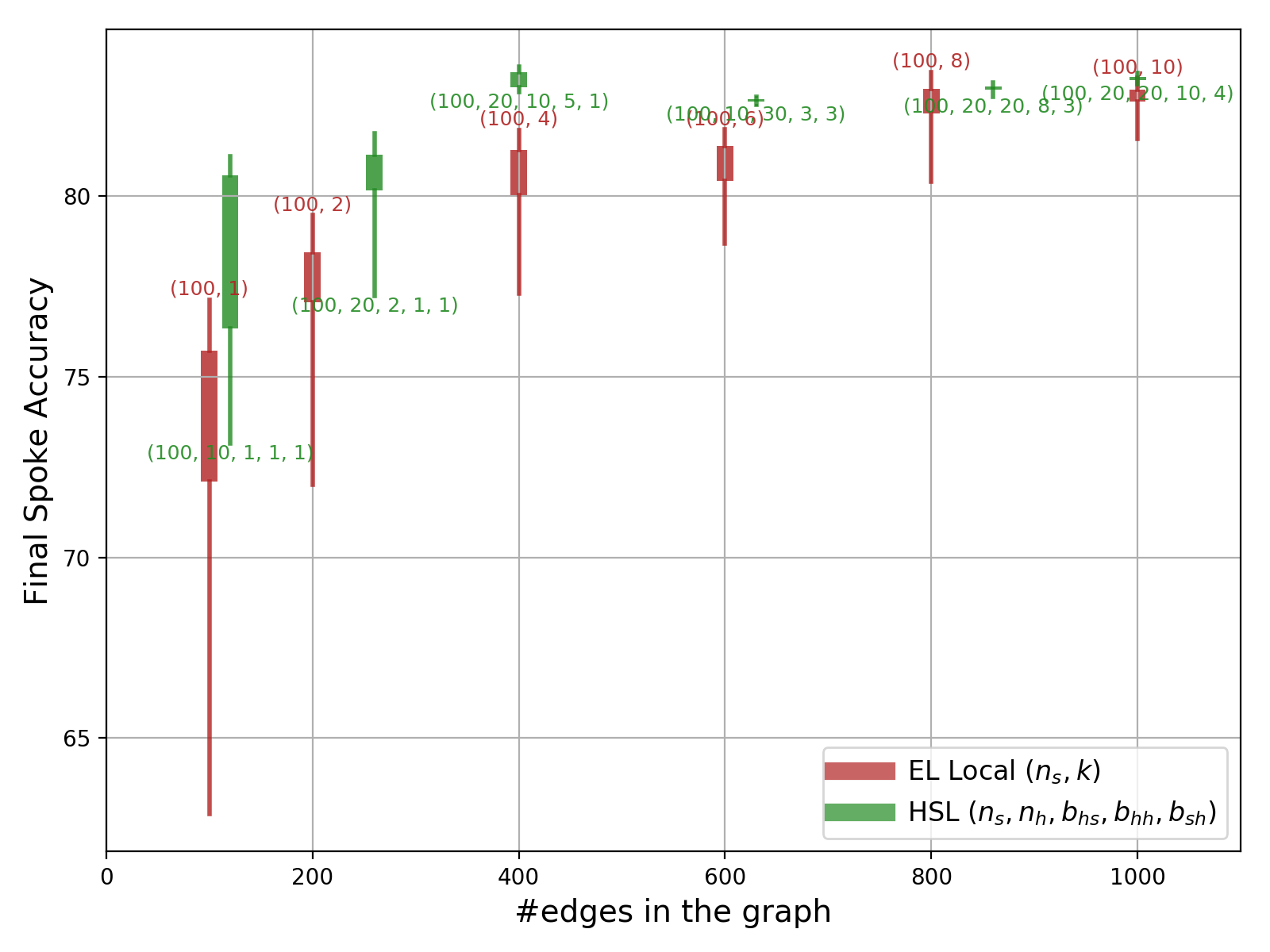}
\hfill
\includegraphics[width=0.48\textwidth]{sections/figures/cdr-s100.png}
\caption{\textbf{HSL vs.\ ELL on AG News (\(n_s=100\)).} 
    The left plot shows the final accuracy distribution, while the right plot presents the consensus distance ratio (CDR). 
    HSL with only 400 edges matches ELL's performance with 1000 edges. 
    The CDR plot continues to confirm the superior mixing efficiency of HSL over ELL.}
\label{fig:agnews-bud-acc-s100}
\end{figure*}

Figure~\ref{fig:bud-acc-cifar} illustrates the final spoke test-accuracy distributions and consensus distance ratios (CDR) for CIFAR-10 with 100 and 200 spokes. 
Similarly, Figures~\ref{fig:agnews-bud-acc-s100} and~\ref{fig:agnews-bud-acc-s200} present these results for AG News. 
We use candle plots to visualize final accuracy, where the candle body represents the interquartile range (25th--75th percentile) and wicks mark the minimum and maximum across the spokes. 
For each configuration shown in the candle plot, a corresponding CDR plot shows the ratio of the post-mixing consensus distance to the pre-mixing distance in each round. 
Since this ratio is typically less than 1, we report its negative logarithm making higher values reflect stronger mixing efficiency.

\paragraph{Experimental Setup:}
To assess the effectiveness of \hsl, we evaluate its performance on two machine learning tasks: image classification on CIFAR-10~\cite{krizhevsky2009learning} and text classification on AG News~\cite{zhang2015character}. In both datasets, data is distributed across \( n_s \) spokes in a non-iid manner using a Dirichlet distribution with \(\alpha=1\). Following our discussion on mixing effectiveness, we compare \hsl\ against \ellocal, monitoring both test accuracy and consensus distance across 500 communication rounds.

For CIFAR-10, each spoke trains a simple CNN with two convolutional layers, a pooling layer, and two fully connected layers (4.2M parameters). Training is done using stochastic gradient descent (SGD) with a constant learning rate of 0.01, a batch size of 128, and three local mini-batch updates per communication round.

For AG News, each spoke trains a lightweight Transformer model with 12.9M parameters, consisting of an embedding layer, two Transformer encoder layers, and a final classification head. The vocabulary size is 95,812, and input sequences are padded to a maximum length of 207 tokens. Training is performed with SGD using a learning rate of 0.05, a batch size of 64, and five local iterations per round.

All experiments were conducted on NVIDIA A100 GPUs, with a maximum memory usage of 40 GB across all configurations. We evaluate multiple configurations of \hsl\ and \ellocal\ and present the results in the following sections.

\paragraph{Comparison with ELL Under Varying Budgets:}

We compare HSL and ELL for 100 and 200 spokes on CIFAR-10 and AG News, varying the total communication budget (the number of directed edges in the graph). After 500 rounds of training, we report each spoke’s final test accuracy against this budget. The total edge count serves as a fair metric because both the system’s per-round communication (number of messages) and computation (volume of model aggregation) scale proportionally with the number of directed edges.

For ELL, the configuration is represented by the tuple \((n_s, k)\), where \(n_s\) is the number of spokes and \(k\) is the outdegree per spoke in each round. The total number of directed edges in this case is given by \(n_s \cdot k\). 

For HSL, the configuration is defined by the tuple \((n_s, n_h, b_{hs}, b_{hh}, b_{sh})\), where \(n_s\) represents the number of spokes, \(n_h\) denotes the number of hubs, and \(b_{hs}\), \(b_{hh}\), and \(b_{sh}\) represent the hub-spoke, hub-hub, and spoke-hub budgets, respectively. The total number of directed edges in HSL is calculated as \(n_h \cdot b_{hs} + n_s \cdot b_{sh} + n_h \cdot b_{hh}\).

We observe that HSL consistently outperforms ELL across various budgets, datasets, and spoke counts (\(n_s = 100, 200\)). 
In Figure~\ref{fig:bud-acc-cifar}, the HSL configuration \((100,5,2,2,2)\) uses just 220 edges \((5\times2 + 5\times2 + 100\times2)\) yet matches the performance of ELL \((100,10)\) with 1000 edges. 
Although a 220-edge HSL graph with 5 hubs is highly sparse, it still maintains strong accuracy through efficient mixing. \hsl's design enables effective information sharing even under tight budget constraints, allowing scalability to larger networks without a proportional rise in communication and computation costs.
Only the extreme low-budget setting \((b_{hs} = b_{hh} = b_{sh} = 1)\) shows a notably lower candle, but such configurations are rarely practical. 
\textit{By contrast, ELL’s accuracy drops more sharply as edge count decreases.} 
With a moderate budget (\(\ge 400\))—letting each hub connect to at least $b_{hs}=10$ spokes (approximately 10\% of all spokes)—\hsl\ surpasses ELL’s best configuration at \(k=10\) under a 1000-edge budget.

For \(n_s = 200\), \hsl's advantage
becomes even more pronounced, highlighting its cost-effectiveness for larger networks. 
While ELL struggles to exceed 58\% accuracy even at 3000 edges, \hsl\ attains comparable accuracy with only 945 edges. 
The CDR plots clarify \hsl's superior mixing properties: except for the minimal 10-hub case \((b_{hs} = b_{hh} = b_{sh} = 1)\), all \hsl\ configurations achieve stronger mixing than ELL’s 1000-edge setup. 
As a result, \hsl\ consistently reduces model variance per round, leading to tighter final accuracy distributions, as reflected in its shorter candle plots.

Even on the AG News dataset, \hsl\ maintains its efficiency, matching ELL’s performance while requiring significantly fewer edges.
—requiring just 400 compared to ELL’s 1000 at $n_s=100$ and only 410 versus 3000 at $n_s=200$, as shown in Figures~\ref{fig:agnews-bud-acc-s100} and~\ref{fig:agnews-bud-acc-s200}.

\begin{figure}
    \centering
    \includegraphics[width=\columnwidth]{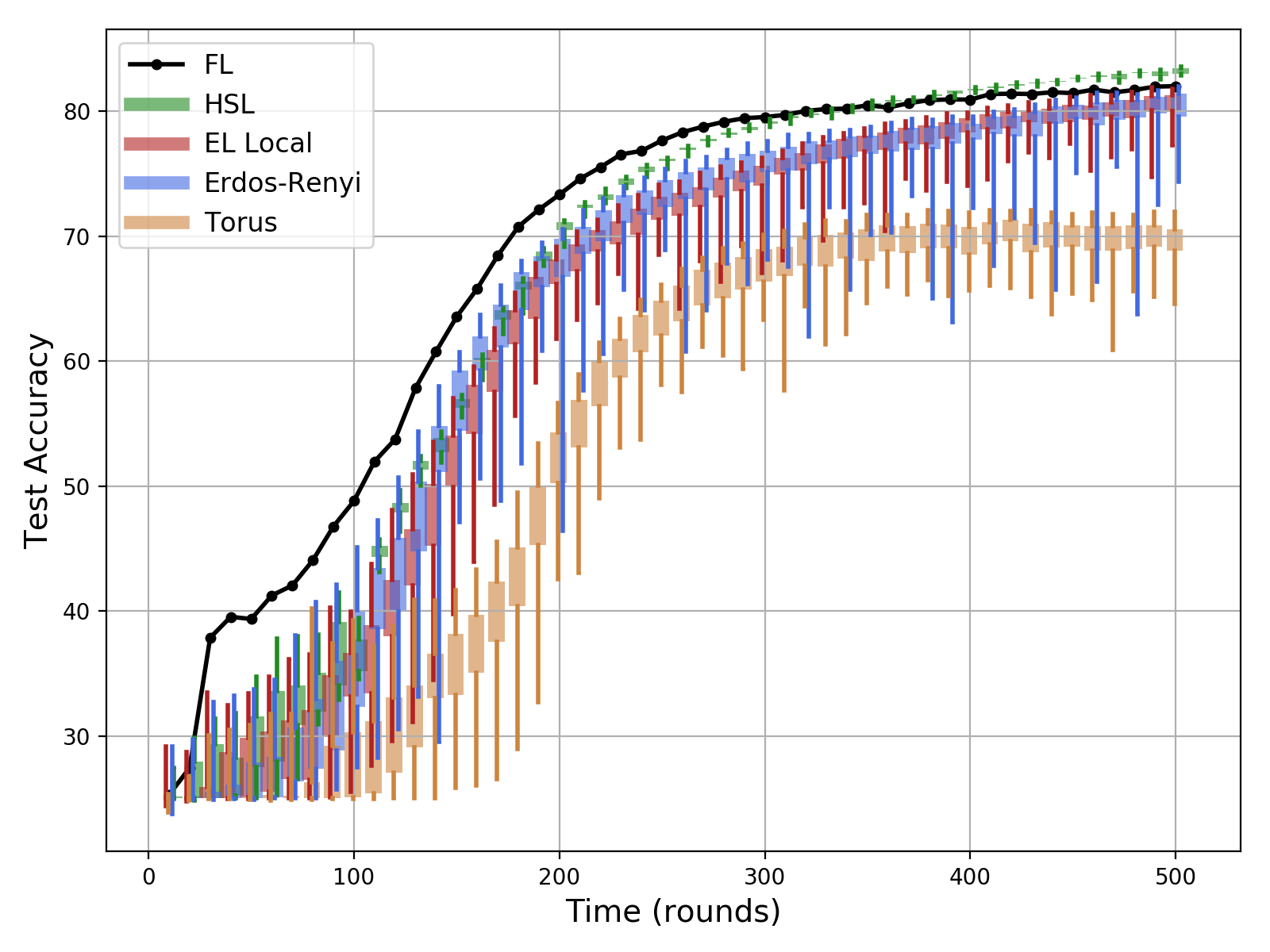}
    \vspace{-25 pt}
    \caption{\textbf{Test Accuracy vs.\ Training Rounds on AG News (\(n_s=100\)).} 
We compare FedAvg (200 edges) with HSL and other decentralized methods (400 edges). 
HSL consistently outperforms decentralized baselines and closely follows FedAvg.
Torus reaches 70\% accuracy, serving as a baseline, while EL Local and Erdős-Rényi exhibit similar trends due to their dynamic graphs.
}
    \label{fig:acc-time}
\end{figure}

\paragraph{Accuracy-Time Plot and Baseline Comparisons}
It is important to note that HSL not only achieves higher final test accuracy but also sustains this advantage throughout training. Figure~\ref{fig:acc-time} shows a typical test accuracy trajectory over training rounds for HSL and ELL at an equal budget of 400 edges for 100 spokes on the AG News dataset. For comparison, we also include FedAvg and two additional decentralized baselines—Torus and Erdős-Rényi\cite{Erdos1984OnTE} both configured with 400 directed edges.

FedAvg achieves the fastest convergence, leveraging exact consensus to ensure ideal mixing even with just 200 edges. This advantage arises from the presence of a centralized server that efficiently coordinates updates.

Torus represents a structured, low-degree topology where each node connects to a fixed set of neighbors in a 2D grid-like pattern. As shown in Figure~\ref{fig:acc-time}, it reaches 70\% accuracy, serving as a baseline for decentralized learning. EL Local and Erdős-Rényi provide competitive alternatives at equal budgets, as both leverage dynamically sampled random graphs for communication. However, they differ in degree distribution—EL Local maintains a fixed, uniform out-degree per node, whereas Erdős-Rényi constructs a random graph with edges assigned independently with a fixed probability.

HSL consistently outperforms the decentralized baselines and closely tracks FedAvg, even briefly surpassing it in the final rounds, as seen in Figure~\ref{fig:acc-time}. This momentary advantage may, however, stem from inherent randomness in decentralized updates.

In summary, these experiments demonstrate that HSL outperforms or closely matches ELL and other decentralized baselines, all while using significantly fewer communication edges. The improved mixing in HSL is evident from higher test accuracy, reduced variance across spokes, and faster convergence in most settings. To further reinforce these findings, we extend our evaluation with a mathematical simulation that directly examines the mixing properties of HSL, ELL, and Erdos-Renyi serving as our baseline.

To further substantiate these findings, we complement our empirical evaluation with a mathematical simulation. Specifically, we analyze the mixing properties of HSL and ELL through a random graph sampling process. At each round, we generate a fresh random graph according to the configuration and compute the effective mixing matrix. The spectral gap of this matrix, defined as the difference between the largest and second-largest eigenvalues of the transition matrix, serves as a widely accepted measure of graph connectivity~\cite{lovasz1993random, chung1997spectral}. A larger spectral gap indicates faster mixing and improved convergence properties in decentralized learning.

\begin{figure}
    \centering
    \includegraphics[width=\columnwidth]{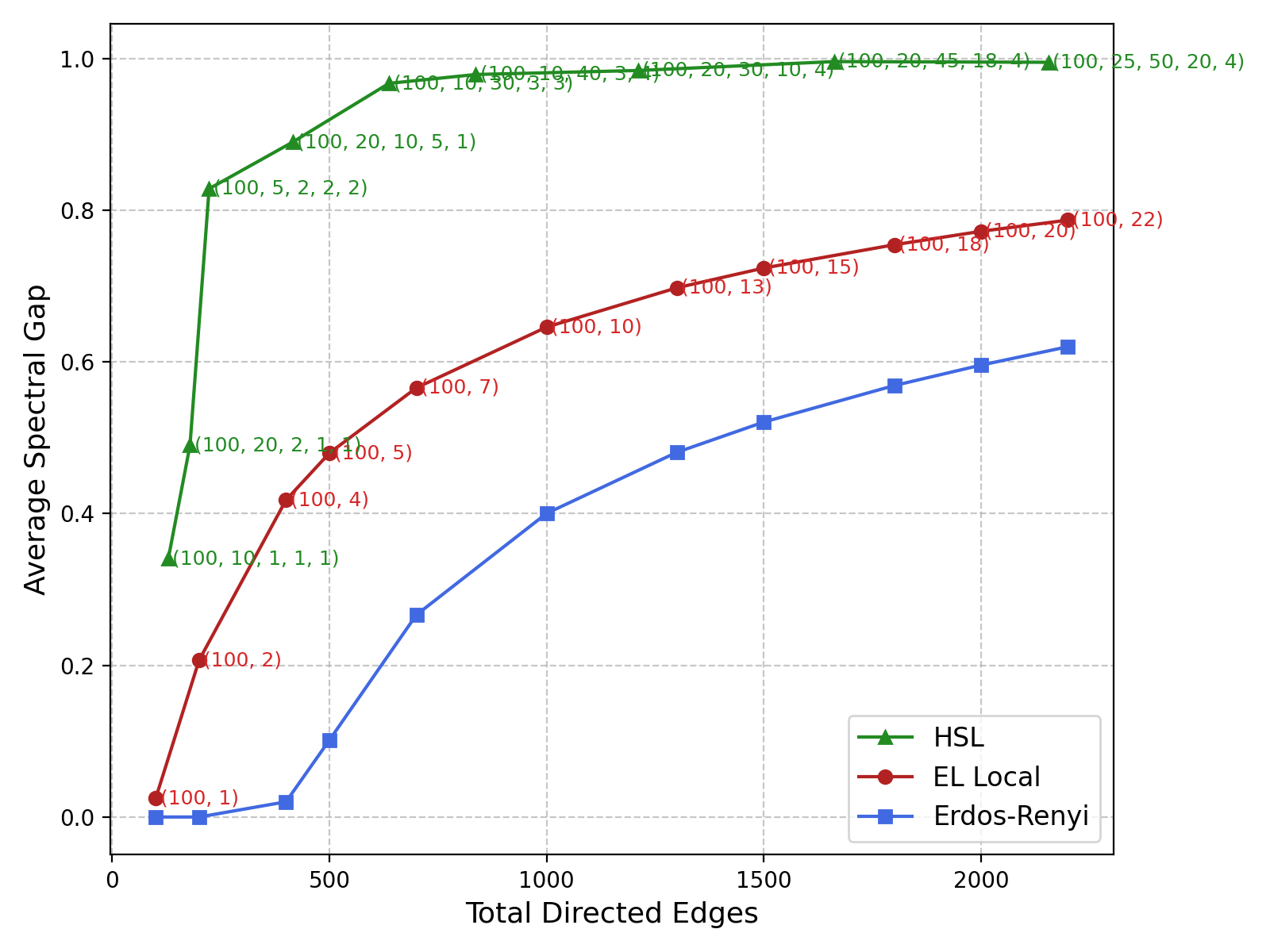}
    \vspace{-25 pt}
    \caption{\textbf{Average spectral gap variation of HSL, ELL, and Erdős-Rényi with total directed edges.} 
The spectral gap was computed from the effective mixing matrix, sampled at each round for 1000 rounds with 100 spokes, and then averaged. 
Erdős-Rényi serves as a reference baseline for comparison. 
The results reaffirm HSL’s superior mixing efficiency, even in mathematical simulations.}

    \label{fig:spectral-gap}
\end{figure}

We repeat this process for 1000 rounds, averaging the spectral gap across all realizations. Figure~\ref{fig:spectral-gap} illustrates the results, conclusively demonstrating that HSL consistently achieves a significantly higher spectral gap than ELL for all comparable budgets between 0 and 2200, where both methods begin to converge. Notably, HSL not only converges faster but also attains a higher final spectral gap value, reinforcing its superior mixing efficiency and scalability.

\vspace{-5 pt}
\section{Discussion}

In this work, we introduced \hsl\ as a scalable and resilient framework that merges the strengths of Federated Learning (FL) and decentralized Peer-to-Peer Learning (P2PL). By structuring communication into hubs and spokes, \hsl\ enables efficient model mixing at the hub level while reducing the communication burden on individual spokes. Our empirical results validate this design: \hsl\ consistently outperforms or matches \ellocal\ while requiring significantly fewer communication edges.

Several key insights emerge from our evaluation. First, the hierarchical structure with decentralized hubs at the top allows \hsl\ to achieve efficient information propagation without the bottlenecks of FL or the increasing connectivity demands of fully decentralized methods. Second, \hsl\ maintains its efficiency as network size scales, demonstrating strong performance even at larger spoke counts. 
Beyond these, two additional advantages stand out. The ability of \hsl\ to deliver strong results even with low budgets and sparse connections suggests future strategies where only a subset of spokes participate in each update round, allowing others to conserve resources while maintaining overall model consistency. Additionally, the \emph{receiver-driven} selection in \hsl\ — where nodes independently choose whom to receive updates from, when hubs and spokes interact— enhances robustness, a property that provides natural resilience against targeted attacks, a promising avenue for future security-focused analyses.

\hsl\ achieves strong mixing at lower budgets, making it well-suited for hybrid systems with both low-power edge devices and high-performance cloud servers. 
Overall, \hsl\ bridges the gap between FL and fully decentralized learning, offering a scalable, resilient, and communication-efficient framework for large-scale collaborative learning systems.

\clearpage
\bibliography{references}
\bibliographystyle{icml2025}

\newpage
\appendix
\onecolumn 

\section{Useful Lemmas}
In this section, we present several key lemmas that are instrumental in establishing the convergence results and analyzing the properties of \hsl.

\begin{lemma}
Average preservation in expectation:  
The average of the models across the network remains preserved in expectation through all stages of the process:
\[
\mathbb{E}\big[\overline{x}_{t+1}\big] = \mathbb{E}\big[\overline{x}_{t+\frac{3}{4}}\big] = \mathbb{E}\big[\overline{x}_{t+\frac{2}{4}}\big] = \mathbb{E}\big[\overline{x}_{t+\frac{1}{4}}\big].
\]
\label{lemma: 1a}
\end{lemma}

\begin{lemma}
For each stage of mixing in \hsl, the consensus distance is recursively bounded as follows:

\begin{enumerate}
    \item \textbf{Spoke-to-Hub Push:}  
    \begin{align*}
    \frac{1}{n_h^2} \sum_{\substack{i, j \in [n_h] \\ i \neq j}} \mathbb{E}\!\left[\left\|x^{(i)}_{t+\tfrac{2}{4}} - x^{(j)}_{t+\tfrac{2}{4}}\right\|^2\right]
    \;\le\;
    \frac{\beta_{hs}}{n_s^2} \sum_{\substack{i, j \in [n_s] \\ i \neq j}} \mathbb{E}\!\left[\left\|x^{(i)}_{t+\tfrac{1}{4}} - x^{(j)}_{t+\tfrac{1}{4}}\right\|^2\right],
    \end{align*}
    where  
    \[
    \beta_{hs} = \frac{1}{b_{hs}} \left[ 1 - \frac{b_{hs} - 1}{n_s - 1} \right].
    \]

    \item \textbf{Hub Gossip:}  
    \begin{align*}
    \frac{1}{n_h^2} \sum_{\substack{i, j \\ i \neq j}} \mathbb{E}\!\left[\left\|x_{t+\tfrac{3}{4}}^{(i)} - x_{t+\tfrac{3}{4}}^{(j)}\right\|^2\right]
    \;\leq\;
    \frac{\beta_{hh}}{n_h^2} \sum_{\substack{i, j \\ i \neq j}} \mathbb{E}\!\left[\left\|x_{t+\tfrac{2}{4}}^{(i)} - x_{t+\tfrac{2}{4}}^{(j)}\right\|^2\right],
    \end{align*}
    where  
    \[
    \beta_{hh} = \frac{1}{b_{hh}} \left( 1 - \left( 1 - \frac{b_{hh}}{n_h - 1} \right)^{n_h} \right) - \frac{1}{n_h - 1}.
    \]

    \item \textbf{Hub-to-Spoke Pull:}  
    \begin{align*}
    \frac{1}{n_s^2} \sum_{\substack{i, j \in [n_s] \\ i \neq j}} \mathbb{E}\!\left[\left\|x^{(i)}_{t+1} - x^{(j)}_{t+1}\right\|^2\right]
    \;\le\;
    \frac{\beta_{sh}}{n_h^2} \sum_{\substack{i, j \in [n_h] \\ i \neq j}} \mathbb{E}\!\left[\left\|x^{(i)}_{t+\tfrac{3}{4}} - x^{(j)}_{t+\tfrac{3}{4}}\right\|^2\right],
    \end{align*}
    where  
    \[
    \beta_{sh} = \frac{1}{b_{sh}} \left[ 1 - \frac{b_{sh} - 1}{n_h - 1} \right].
    \]

    \item \textbf{Final Consensus Bound:}  
    Combining the above three stages, the consensus distance at \(x_{t+1}\) is bounded in terms of the distance at \(x_{t+\frac{1}{4}}\):
    \begin{align*}
    \frac{1}{n_s^2} \sum_{\substack{i, j \in [n_s] \\ i \neq j}} \mathbb{E}\!\left[\left\|x^{(i)}_{t+1} - x^{(j)}_{t+1}\right\|^2\right]
    \;\le\;
    \beta_{HSL} \cdot \frac{1}{n_s^2} \sum_{\substack{i, j \in [n_s] \\ i \neq j}} \mathbb{E}\!\left[\left\|x^{(i)}_{t+\tfrac{1}{4}} - x^{(j)}_{t+\tfrac{1}{4}}\right\|^2\right],
    \end{align*}
    where  
    \[
    \beta_{HSL} = \beta_{hs} \cdot \beta_{hh} \cdot \beta_{sh}.
    \]
\end{enumerate}

\label{lemma: 1b}
\end{lemma}

\begin{lemma}
For each stage of aggregation in \hsl, the expected deviation of the average model is bounded as follows:

\begin{enumerate}
    \item \textbf{Spoke-to-Hub Push:}  
    \begin{align*}
        \mathbb{E} \left[ \left\| \bar{x}_{t+\frac{2}{4}} - \bar{x}_{t+\frac{1}{4}} \right\|^2 \right] 
        = \frac{\beta_{hs}}{n_s n_h} \sum_{i \in [n_s]} \mathbb{E} \left[ \left\| x_{t+\frac{1}{4}}^{(i)} - \bar{x}_{t+\frac{1}{4}} \right\|^2 \right].
    \end{align*}

    \item \textbf{Hub Gossip:}  
    \begin{align*}
        \mathbb{E} \left[\left\|x_{t+\frac{3}{4}} - \bar{x}_{t+\frac{2}{4}}\right\|^2\right] 
        \leq \frac{\beta_{hh}}{n_h^2} \sum_{i \in [n_h]} \mathbb{E} \left[\left\|x^{(i)}_{t+\frac{2}{4}} - \bar{x}_{t+\frac{2}{4}}\right\|^2\right].
    \end{align*}

    \item \textbf{Hub-to-Spoke Pull:}  
    \begin{align*}
        \mathbb{E} \left[ \left\| \bar{x}_{t+1} - \bar{x}_{t+\frac{3}{4}} \right\|^2 \right] 
        = \frac{\beta_{sh}}{n_s n_h} \sum_{i \in [n_h]} \mathbb{E} \left[ \left\| x_{t+\frac{3}{4}}^{(i)} - \bar{x}_{t+\frac{3}{4}} \right\|^2 \right].
    \end{align*}
\end{enumerate}

\label{lemma: 1c}
\end{lemma}

\begin{lemma}
The expected consensus distance and gradient variance across spokes are bounded as follows:

\begin{enumerate}
    \item \textbf{Consensus Distance Bound:}  
    \begin{align*}
        \frac{1}{n_s^2} \sum_{i,j \in [n_s]} \mathbb{E} \left[ \left\| x_t^{(i)} - x_t^{(j)} \right\|^2 \right] 
        \leq 20 \frac{1 + 3\beta_{HSL}}{(1 - \beta_{HSL})^2} \beta_{HSL} \gamma^2 (\sigma^2 + \mathcal{H}^2).
    \end{align*}

    \item \textbf{Gradient Variance Bound:}  
    \begin{align*}
        \frac{1}{n_s^2} \sum_{i,j \in [n_s]} \mathbb{E} \left[ \left\| g_t^{(i)} - g_t^{(j)} \right\|^2 \right] 
        \leq 15 (\sigma^2 + \mathcal{H}^2).
    \end{align*}
\end{enumerate}

\label{lemma: 2}
\end{lemma}

\begin{lemma}
The expected gradient norm of the global objective satisfies the following upper bound:
\begin{align*}
    \mathbb{E}\left[\left\|\nabla F(\bar{x}_t)\right\|^2\right] 
    &\leq \frac{2}{\gamma}\mathbb{E}\left[F(\bar{x}_t) - F(\bar{x}_{t+1})\right] 
    + \frac{L}{2 n^2}\sum_{i,j} \mathbb{E}\left[\left\|x_t^{(i)} - x_t^{(j)}\right\|^2\right] 
    + \frac{4L\gamma \sigma^2}{n} \\
    &\quad + \frac{4L}{\gamma} \mathbb{E} \Big[ \left\|\bar{x}_{t+1} - \bar{x}_{t+\frac{3}{4}}\right\|^2 
    + \left\|\bar{x}_{t+\frac{3}{4}} - \bar{x}_{t+\frac{2}{4}}\right\|^2 
    + \left\|\bar{x}_{t+\frac{2}{4}} - \bar{x}_{t+\frac{1}{4}}\right\|^2 \Big].
\end{align*}
\label{lemma: 3}
\end{lemma}

\begin{lemma}
Variance decomposition:  
For any set of vectors \(\{x^{(i)}_t, i \in [n_s]\}\),
\[
\frac{1}{n_s}\sum_{i} \left\|x^{(i)}_t - \bar{x}_t\right\|^2
\;=\;
\frac{1}{2}\,\frac{1}{n_s^2}\sum_{\substack{i, j \\ i \neq j}} \left\|x^{(i)}_t - x^{(j)}_t\right\|^2.
\]
\label{lemma-5}
\end{lemma}

\section{Proof of Theorem 1}
\begin{proof}[Proof]
Recall that for any vectors \(\mathbf{a}, \mathbf{b} \in \mathbb{R}^d\), Jensen's inequality (for the \(\ell_2\)-norm) states:
\[
\left\|\mathbf{a} + \mathbf{b}\right\|^2 \;\le\; 2\,\left\|\mathbf{a}\right\|^2 \;+\; 2\,\left\|\mathbf{b}\right\|^2.
\]
We apply this inequality with \(\mathbf{a} = \nabla F(\bar{x_t})\) and 
\(\mathbf{b} = \nabla F(x_t^{(i)}) - \nabla F(\bar{x_t})\). For any \(i \in [n_s]\), we obtain
\begin{align*}
\mathbb{E} \left[\left\|\nabla F(x_t^{(i)})\right\|^2\right] 
&= \mathbb{E} \left[\left\|\nabla F(\bar{x_t}) + \left(\nabla F(x_t^{(i)}) - \nabla F(\bar{x_t})\right)\right\|^2\right] \\
&\le 2\,\mathbb{E} \left[\left\|\nabla F(\bar{x_t})\right\|^2\right] 
\;+\; 2\,\mathbb{E} \left[\left\|\nabla F(x_t^{(i)}) - \nabla F(\bar{x_t})\right\|^2\right].
\end{align*}
Using Assumption~\ref{ass:smoothness} (\emph{Smoothness}), which implies 
\(\left\|\nabla F(x) - \nabla F(y)\right\|\le L\,\left\|x-y\right\|\), we further bound the second term to obtain:
\begin{align*}
\mathbb{E} \left[\left\|\nabla F(x_t^{(i)})\right\|^2\right] 
&\le 2\,\mathbb{E} \left[\left\|\nabla F(\bar{x_t})\right\|^2\right] 
\;+\; 2\,L^2\mathbb{E}\left[\left\|x_t^{(i)} - \bar{x_t}\right\|^2\right].
\end{align*}

Next, we average over all \(i \in [n_s]\):
\[
\frac{1}{n_s} \sum_{i=1}^{n_s} \mathbb{E}\left[\left\|\nabla F(x_t^{(i)})\right\|^2\right] 
\;\le\; 2\,\mathbb{E}\left[\left\|\nabla F(\bar{x_t})\right\|^2\right] 
\;+\; \frac{2L^2}{n_s} \sum_{i=1}^{n_s} \mathbb{E}\left[\left\|x_t^{(i)} - \bar{x_t}\right\|^2\right].
\]

Finally, making use of Lemma~\ref{lemma-5}, which states
\[
\frac{1}{n_s}\sum_{i=1}^{n_s}\left\|x_t^{(i)} - \bar{x_t}\right\|^2
= \frac{1}{2n_s^2}\sum_{i,j\in[n_s]}\left\|x_t^{(i)} - x_t^{(j)}\right\|^2,
\]
we get
\begin{align*}
\frac{1}{n_s} \sum_{i=1}^{n_s} \mathbb{E} \left[\left\|\nabla F(x_t^{(i)})\right\|^2\right] 
&\le 2\,\mathbb{E} \left[\left\|\nabla F(\bar{x_t})\right\|^2\right]
\;+\; \frac{L^2}{n_s^2} \sum_{i,j\in[n_s]} \mathbb{E} \left[\left\|x_t^{(i)} - x_t^{(j)}\right\|^2\right].
\end{align*}

Bounding the first term on the RHS using Lemma~\ref{lemma: 3}, we further obtain: 
\begin{align}
\frac{1}{n_s} \sum_{i=1}^{n_s} \mathbb{E} \left[\left\|\nabla F(x_t^{(i)})\right\|^2\right] 
&\leq \frac{4}{\gamma}\,\mathbb{E} \left[F(\bar{x}_t) \;-\; F(\bar{x}_{t+1})\right]
\;+\; \frac{2L^2}{n_s^2} \sum_{i,j \in [n_s]} \mathbb{E} \left[\left\|x_t^{(i)} - x_t^{(j)}\right\|^2\right] \nonumber \\
&\quad+\; \frac{8L\gamma \sigma^2}{n_s} 
\;+\; \frac{8L}{\gamma}\,\mathbb{E} \left[\left\|\bar{x}_{t+1} - \bar{x}_{t+\frac{3}{4}}\right\|^2 
\;+\; \left\|\bar{x}_{t+\frac{3}{4}} - \bar{x}_{t+\frac{2}{4}}\right\|^2 
\;+\; \left\|\bar{x}_{t+\frac{2}{4}} - \bar{x}_{t+\frac{1}{4}}\right\|^2\right].
\label{eq: 6}
\end{align}

Using Lemma~\ref{lemma: 1b}, we also have:

\begin{align*}
    \mathbb{E} \left[ \left\| \bar{x}_{t+1} - \bar{x}_{t+\frac{3}{4}} \right\|^2\right] 
    &\leq \frac{\beta_{sh} \beta_{hh} \beta_{hs}}{2n_s^3} \sum_{i,j \in [n_s]} \mathbb{E} \left[ \left\| x_{t+\frac{1}{4}}^{(i)} - x_{t+\frac{1}{4}}^{(j)} \right\|^2 \right]\\
    \mathbb{E} \left[ \left\| \bar{x}_{t+\frac{3}{4}} - \bar{x}_{t+\frac{2}{4}} \right\|^2\right] 
    &\leq \frac{\beta_{hh} \beta_{hs}}{2n_h n_s^2} \sum_{i,j \in [n_s]} \mathbb{E} \left[ \left\| x_{t+\frac{1}{4}}^{(i)} - x_{t+\frac{1}{4}}^{(j)} \right\|^2 \right]\\
    \mathbb{E} \left[ \left\| \bar{x}_{t+\frac{2}{4}} - \bar{x}_{t+\frac{1}{4}} \right\|^2\right] 
    &\leq \frac{\beta_{hs}}{2n_h n_s^2} \sum_{i,j \in [n_s]} \mathbb{E} \left[ \left\| x_{t+\frac{1}{4}}^{(i)} - x_{t+\frac{1}{4}}^{(j)} \right\|^2 \right]
\end{align*}

Adding the above inequalites,
\begin{align}
    \mathbb{E} \left[ \left\| \bar{x}_{t+1} - \bar{x}_{t+\frac{3}{4}} \right\|^2\right] 
    \;+\;
    \mathbb{E} \left[ \left\| \bar{x}_{t+\frac{3}{4}} - \bar{x}_{t+\frac{2}{4}} \right\|^2\right] 
    \;+\;
    \mathbb{E} \left[ \left\| \bar{x}_{t+\frac{2}{4}} - \bar{x}_{t+\frac{1}{4}} \right\|^2\right] 
    &\;\;\le\;\;
    \frac{\beta'}{n_s} \frac{1}{n_s^2}
    \sum_{i,j \in [n_s]}
    \mathbb{E} \left[ \left\| x_{t+\frac{1}{4}}^{(i)} - x_{t+\frac{1}{4}}^{(j)} \right\|^2 \right]
\label{eq: beta_dash}
\end{align}
where 
\[
\frac{\beta'}{n_s} \;=\;
\frac{\beta_{hs}}{2n_h} + \frac{\beta_{hh} \beta_{hs}}{2n_h} + \frac{\beta_{sh} \beta_{hh} \beta_{hs}}{2n_s}
\]

Remember the partial update step \(x_{t+\tfrac{1}{4}}^{(i)} \triangleq x_t^{(i)} - \gamma\,g_t^{(i)}\). Thus,

\begin{align}
    \mathbb{E}\left[\left\|x_{t+\tfrac{1}{4}}^{(i)} - x_{t+\tfrac{1}{4}}^{(j)}\right\|^2\right] 
    &= \mathbb{E}\left[\left\|x_t^{(i)} - \gamma\,g_t^{(i)} \;-\; x_t^{(j)} + \gamma\,g_t^{(j)}\right\|^2\right] \nonumber \\
    &\le 2\,\mathbb{E}\left[\left\|x_t^{(i)} - x_t^{(j)}\right\|^2\right]
    \;+\;
    2\,\gamma^2\,\mathbb{E}\left[\left\|g_t^{(i)} - g_t^{(j)}\right\|^2\right].
\label{eq: g}
\end{align}
where we make use of Young's inequality.

Substituting ~\ref{eq: g} and ~\ref{eq: beta_dash} into ~\ref{eq: 6}, we get:

\begin{align}
    \frac{1}{n_s} \sum_{i=1}^{n_s} \mathbb{E} \left[ \left\| \nabla F(x_t^{(i)}) \right\|^2 \right] 
    &\leq \frac{4}{\gamma} \mathbb{E} \left[ F(\bar{x}_t) - F(\bar{x}_{t+1}) \right] + \frac{8L \gamma \sigma^2}{n_s} \nonumber \\
    & + \left({2L^2} + \frac{16L \beta'}{\gamma n_s}\right) \frac{1}{n_s^2} \sum_{i=1}^{n_s} \mathbb{E} \left[ \left\| x_{t}^{(i)} - x_{t}^{(j)} \right\|^2 \right] +  \frac{16L \gamma \beta'}{n_s} \frac{1}{n_s^2} \sum_{i=1}^{n_s} \mathbb{E} \left[ \left\| g_{t}^{(i)} - g_{t}^{(j)} \right\|^2 \right]
\label{eq: 12}
\end{align}

From Remark ~\ref{remark: 2}, we have: $\beta_{HSL} \leq 1-\frac{1}{e}$
Therefore,
\begin{align*}
20 \frac{1+3\beta_{HSL}}{(1-\beta_{HSL})^2} \leq 500
\end{align*}
We substitute this in~\cref{lemma: 2} to get
\begin{align*}
    \frac{1}{n_s^2} \sum_{i=1}^{n_s} \mathbb{E} \left[ \left\| x_{t}^{(i)} - x_{t}^{(j)} \right\|^2 \right] 
    &\leq 500 \beta_{HSL} \gamma^2 (\sigma^2 + \mathcal{H}^2)
\end{align*}
From~\cref{lemma: 2}, we also have,
\begin{align*}
    \frac{1}{n_s^2} \sum_{i=1}^{n_s} \mathbb{E} \left[ \left\| g_{t}^{(i)} - g_{t}^{(j)} \right\|^2 \right] 
    &\leq 15 (\sigma^2 + \mathcal{H}^2)\\
\end{align*}

Substituting this in ~\ref{eq: 12}, we obtain:
\begin{align*}
    \frac{1}{n_s} \sum_{i=1}^{n_s} \mathbb{E} \left[ \left\| \nabla F(x_t^{(i)}) \right\|^2 \right] 
    &\leq \frac{4}{\gamma} \mathbb{E} \left[ F(\bar{x}_t) - F(\bar{x}_{t+1}) \right] + \left({2L^2}  + \frac{16 L \beta'}{\gamma n_s} \right) 500 \beta_{HSL} \gamma^2 (\sigma^2 + \mathcal{H}^2) + \frac{8L\gamma\sigma^2}{n_s} + \frac{240}{n_s}L \gamma \beta_{HSL} (\sigma^2 + \mathcal{H}^2)
\end{align*}
Taking the average over $t \in {0,...,T-1}$, we obtain:
\begin{align}
    \frac{1}{n_s T} \sum_{t=0}^{T-1} \sum_{i=1}^{n_s} \mathbb{E} \left[ \left\| \nabla F(x_t^{(i)}) \right\|^2 \right] 
    &\leq \frac{4}{T \gamma} \Delta_0 + \frac{\gamma}{n_s} \left(16 L \beta' 500 \beta_{HSL} (\sigma^2 + \mathcal{H}^2) + 8L\sigma^2 + 240L \beta_{HSL} (\sigma^2 + \mathcal{H}^2) \right) \nonumber \\
    &\quad \quad \quad \quad + \gamma^2 \left(2L^2 500 \beta_{HSL} (\sigma^2 + \mathcal{H}^2) \right) \nonumber \\
    &\leq \frac{4}{T \gamma} \Delta_0 + \frac{8L\gamma}{n_s} \left((1+663 \beta')\sigma^2 + 663\beta' \mathcal{H}^2\right) + \gamma^2 \left(1000L^2\beta_{HSL}(\sigma^2+\mathcal{H}^2) \right)
\label{eq: 15}
\end{align}
Here, we make use of the fact that since
$\beta_{HSL} \leq 1-\frac{1}{e}$, 
$1000\beta_{HSL} \leq 663$
Now, setting 
\begin{align}
\gamma = min \left\{\sqrt{\frac{n_s \Delta_0}{2TL((1+663 \beta')\sigma^2 + 663\beta' \mathcal{H}^2)}}, \sqrt[3]{\frac{\Delta_0}{250TL^2 \beta_{HSL}(\sigma^2+\mathcal{H}^2)}}, \frac{1}{20L}\right\}
\label{eq: 16}
\end{align}
we have
\begin{align}
\frac{1}{\gamma} &= max\left\{\sqrt{\frac{2TL((1+663 \beta')\sigma^2 + 663\beta' \mathcal{H}^2)}{n_s \Delta_0}}, \sqrt[3]{\frac{250TL^2 \beta_{HSL}(\sigma^2+\mathcal{H}^2)}{\Delta_0}}, 20L \right\} \nonumber \\
& \leq \sqrt{\frac{2TL((1+663 \beta')\sigma^2 + 663\beta' \mathcal{H}^2)}{n_s \Delta_0}} + \sqrt[3]{\frac{250TL^2 \beta_{HSL}(\sigma^2+\mathcal{H}^2)}{\Delta_0}} + 20L
\label{eq: 17}
\end{align}
Plugging equation~\ref{eq: 17} and equation~\ref{eq: 16} in equation~\ref{eq: 15} we obtain
\begin{align*}
    \frac{1}{n_s T} \sum_{t=0}^{T-1} \sum_{i=1}^{n_s} \mathbb{E} \left[ \left\| \nabla F(x_t^{(i)}) \right\|^2 \right] 
    &\leq 8 \sqrt{\frac{2L \Delta_0}{T n_s} (1+663 \beta')\sigma^2 + 663\beta' \mathcal{H}^2} + 51\sqrt[3]{\frac{L^2 \beta_{HSL} \Delta_0^2 (\sigma^2 + \mathcal{H}^2)}{T^2}} + \frac{80L \Delta_0}{T} \\
    &\in \mathcal{O}\left(\sqrt{\frac{L \Delta_0}{T n_s} ((1+\beta')\sigma^2 +\beta'\mathcal{H}^2)} + \sqrt[3]{\frac{L^2 \beta_{HSL} \Delta_0^2 (\sigma^2 + \mathcal{H}^2)}{T^2}} + \frac{L \Delta_0}{T}\right)
\end{align*}
Here, we use the simplification that $\frac{2000}{250^{\frac{2}{3}}} < 51$. \\
This completes the derivation of the stated bound.
\end{proof}
\section{Additional Results and Remarks}

\begin{figure*}[ht]
\centering
\includegraphics[width=0.48\textwidth]{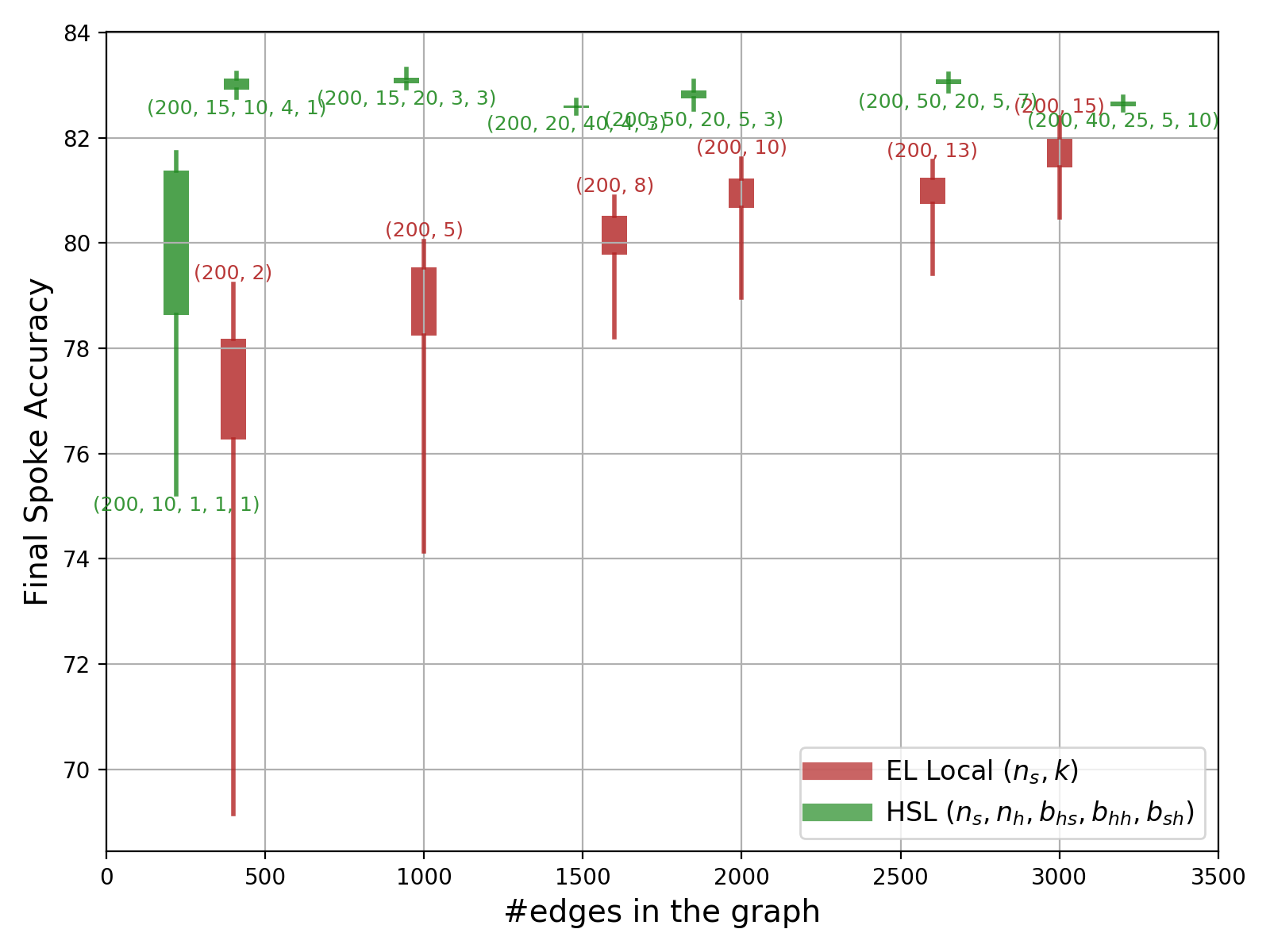}
\hfill
\includegraphics[width=0.48\textwidth]{sections/figures/cdr-s200.png}
\caption{\textbf{HSL vs.\ ELL on AG News (\(n_s=200\)).} 
    Final accuracy distribution on the left and consensus distance ratio (CDR) on the right. 
    HSL with only 410 edges matches ELL's performance with 3000 edges. 
    The CDR plot continues to confirm the superior mixing efficiency of HSL over ELL}
\label{fig:agnews-bud-acc-s200}
\end{figure*}

\begin{remark} Note that $\beta_{hh}$, computed in~\cref{lemma: 1b} is decreasing in $b_{hh}$ and increasing in $n_h$, therefore, for any $b_{hh} \ge 1$, and $n_h \ge 2$ we have
\begin{align*}
\beta_{hh} \Big\vert_{n < \infty} &\le \lim_{n \to \infty} \beta_1 \\
&= \lim_{n \to \infty} \left( 1 - \left( 1 - \frac{1}{n-1} \right)^n - \frac{1}{n-1} \right) \\
&= 1 - \frac{1}{e},
\end{align*}
where $\beta_1$ is the $\beta$ at $b_{hh}=1$. $e$ is Euler's Number and we used the fact that $\lim_{n \to \infty} \left( 1 - \frac{1}{n} \right)^n = \frac{1}{e}$. \\
We also have $\beta_{hs} \leq 1$ and $\beta_{sh} \leq 1$. Multiplying these, we get $\beta_{HSL} \leq 1-\frac{1}{e}$
\label{remark: 2}
\end{remark}

\begin{remark}

\begin{align}
\beta_{hs} &= \frac{1}{b_{hs}} \left( 1 - \frac{b_{hs} - 1}{n_s - 1} \right) \label{eqn:b_hs_1}\\
\beta_{hh} &\le 1 - \frac{1}{e} 
\label{eqn:b_hh_1}\\
\beta_{hs} &\le \frac{1}{b_{sh}} \left( 1 - \frac{b_{hs} - 1}{n_h - 1} \right)
\label{eqn:b_sh_1}
\end{align}
By combining (\ref{eqn:b_hs_1}) and (\ref{eqn:b_sh_1}),
\begin{align*}
\beta_{hs} \beta_{sh} &\le \frac{1}{b_{hs} b_{sh}}\frac{n_s-b_{hs}}{n_s-1} \frac{n_h-b_hs}{n_h - 1} \\
&= \frac{1}{b_{hs} b_{sh}}\frac{(1-\frac{b_{hs}}{n_s})}{(1-\frac{1}{n_s})} \frac{(1-\frac{b_{sh}}{n_h})}{(1 - \frac{1}{n_h})} 
\end{align*}
If we have $b_{sh} n_s \leq b_{hs} n_h$
\begin{align*}
\beta_{hs} \beta_{sh} &\leq \frac{n_s}{b_{hs} b_{sh} n_h}\frac{(1-\frac{b_{hs}}{n_s})}{(1-\frac{1}{n_s})} \frac{(1-\frac{b_{hs}}{n_s})}{(1 - \frac{1}{n_h})}\\
&= \frac{n_s}{n_h} \frac{\left( \frac{1}{b_{hs}} - \frac{1}{n_s} \right)^2}{\left( 1 - \frac{1}{n_s} \right) \left( 1 - \frac{1}{n_h} \right)}
\end{align*}

This is decreasing in  $b_{hs}$. For  $b_{sh} = 1$, (lowest spoke budget), $\beta_{hs} = \frac{n_s}{n_h}$
\begin{align*}
\beta_{hs} \beta_{sh} &\leq \frac{n_s}{n_h} \frac{(n_h-1)^2}{n_s^2} \frac{n_s}{n_s-1}\frac{n_h}{n_h-1}\\
&= \frac{n_h - 1}{n_s - 1} < \frac{n_e}{n_s} \\
\beta_{HSL} &\le \frac{n_h}{n_s} \left( 1 - \frac{1}{e} \right)
\end{align*}

Therefore, we guarantee lower upper bound on $\beta_{HSL}$ as compared to $\beta_{EL}$ under the condition where $n_h \cdot b_{hs} \geq n_s \cdot b_{sh}$
\label{BHSL}
\end{remark}
\section{Proof of Lemmas}
\subsection{Proof of~\cref{lemma: 1a}} 
Here, we prove the \textit{average preservation in expectation} property of \hsl.  
\[
\mathbb{E}\big[\overline{x}_{t+1}\big] = \mathbb{E}\big[\overline{x}_{t+\frac{3}{4}}\big] = \mathbb{E}\big[\overline{x}_{t+\frac{2}{4}}\big] = \mathbb{E}\big[\overline{x}_{t+\frac{1}{4}}\big].
\]

\begin{proof}
From the system dynamics, we have:
\[
X_{t+\frac{2}{4}} = W_{hs} X_{t+\frac{1}{4}},
\]
where \(W_{hs}\) is independent of \(X_{t+\frac{1}{4}}\). Taking expectations:
\[
\mathbb{E}\big[X_{t+\frac{2}{4}}\big] = \mathbb{E}\big[W_{hs}\big] \mathbb{E}\big[X_{t+\frac{1}{4}}\big].
\]

By the construction of \(W_{hs}\) as row-stochastic, we have:
\[
\sum_{j=1}^{n_s} W_{hs}^{(i,j)} = 1 \quad \text{for all hubs } i.
\]
Taking expectations and using symmetry (equal probability for all elements), let \(c = \mathbb{E}\big[W_{hs}^{(i,j)}\big]\). Then:
\[
\sum_{j=1}^{n_s} \mathbb{E}\big[W_{hs}^{(i,j)}\big] = 1 \implies n_s c = 1 \implies c = \frac{1}{n_s}.
\]
Thus:
\[
\mathbb{E}[W_{hs}] = \frac{1}{n_s} \mathbf{1}_{n_h} \mathbf{1}_{n_s}^T,
\]
where \(\mathbf{1}_{n_h}\) and \(\mathbf{1}_{n_s}\) are column vectors of ones of dimension \(n_h\) and \(n_s\), respectively.

Substituting, we get:
\begin{align*}
\mathbb{E}\big[X_{t+\frac{2}{4}}\big] &= \frac{1}{n_s} \mathbf{1}_{n_h} \mathbf{1}_{n_s}^T \mathbf{1}_{n_s} \overline{x}_{t+\frac{1}{4}} \\
&= \mathbf{1}_{n_h} \overline{x}_{t+\frac{1}{4}} \\
&= \mathbb{E}\big[X_{t+\frac{1}{4}}\big]
\end{align*}

Here we use the fact that \(\mathbf{1}_{n_s}^T \mathbf{1}_{n_s} = n_s\).

Thus, \(\mathbb{E}\big[\overline{x}_{t+\frac{2}{4}}\big] = \mathbb{E}\big[\overline{x}_{t+\frac{1}{4}}\big]\)

Now, consider the \textbf{second stage of aggregation} post hub gossip. 
\[
X_{t+\frac{3}{4}} = W_{h} X_{t+\frac{2}{4}},
\]

Let \(n_h\) be the total number of hubs, and let \(A^{(i)} = |\mathcal{A}_k|\) denote the in-degree of the \(i\)-th hub, where the outdegree of every hub is fixed to \(b_{hs}\).
For any hub \(i \in [n_h]\), define \(I_j^{(i)}\) as the indicator function denoting whether the \(j\)-th hub is connected to hub~\(i\).  Then we claim:
\[
\mathbb{E}\big[x_{t+\frac{3}{4}}^{(i)}\big]
\;=\;
\mathbb{E}\Biggl[
  \frac{1}{A^{(i)} + 1}
  \Bigl(
    x_{t+\frac{2}{4}}^{(i)}
    \;+\;
    \sum_{j \in [n_h]\setminus\{i\}} \mathcal{I}_j^{(i)} \,x_{t+\frac{2}{4}}^{(j)}
  \Bigr)
\Biggr].
\]
First, we take a conditional expectation on \(A^{(i)}\):
\begin{align*}
\mathbb{E}\big[x_{t+\frac{3}{4}}^{(i)}\big]
&=\;
\mathbb{E}\Bigl[
   \mathbb{E}\Bigl[
     \frac{1}{A^{(i)} + 1}
     \Bigl(
       x_{t+\frac{2}{4}}^{(i)}
       \;+\;
       \sum_{j \in [n_h]\setminus\{i\}} \mathcal{I}_j^{(i)}\,x_{t+\frac{2}{4}}^{(j)}
     \Bigr)
   \Bigm|\,
     A^{(i)}
   \Bigr]\Bigr]
\\[6pt]
&=\;
\mathbb{E}\Bigl[
  \frac{1}{A^{(i)}+1}
  \Bigl(
    x_{t+\frac{2}{4}}^{(i)}
    \;+\;
    \sum_{j \in [n_h]\setminus\{i\}}
      \mathbb{E}[\mathcal{I}_j^{(i)} \mid A^{(i)}]
      \,x_{t+\frac{2}{4}}^{(j)}
  \Bigr)
\Bigr].
\end{align*}
Since each of the other \(n_h - 1\) hubs has the same probability of sending its value to hub~\(i\), we have
\[
\mathbb{E}\big[I_j^{(i)} \mid A^{(i)}\big]
\;=\;
\frac{A^{(i)}}{n_h - 1}.
\]
Thus,
\begin{align*}
\mathbb{E}\big[x_{t+\frac{3}{4}}^{(i)}\big]
&=\;
\mathbb{E}\Bigl[
  \frac{1}{A^{(i)} + 1}
  \Bigl(
    x_{t+\frac{2}{4}}^{(i)}
    \;+\;
    \frac{A^{(i)}}{n_h - 1}
    \sum_{j \in [n_h]\setminus\{i\}}
      x_{t+\frac{2}{4}}^{(j)}
  \Bigr)
\Bigr]
\\[4pt]
&=\;
\mathbb{E}\Bigl[
  \frac{1}{A^{(i)} + 1}
  \Bigl(
    x_{t+\frac{2}{4}}^{(i)}
    \;+\;
    \frac{A^{(i)}}{n_h - 1}\Bigl(\,n_h\,\bar{x}_{t+\frac{2}{4}} - x_{t+\frac{2}{4}}^{(i)}\Bigr)
  \Bigr)
\Bigr],
\end{align*}
where \(\bar{x}_{t+\frac{2}{4}} = \frac{1}{n_h}\sum_{j=1}^{n_h} x_{t+\frac{2}{4}}^{(j)}\).  Let
\[
p
\;=\;
\mathbb{E}\Bigl[\frac{A^{(i)}}{A^{(i)} + 1}\Bigr].
\]
Collecting terms, it follows that
\[
\mathbb{E}\big[x_{t+\frac{3}{4}}^{(i)}\big]
\;=\;
\frac{p\,n_h}{\,n_h - 1\,}\,\bar{x}_{t+\frac{2}{4}}
\;+\;
\Bigl(1 - \frac{p\,n_h}{\,n_h - 1\,}\Bigr)\,x_{t+\frac{2}{4}}^{(i)}.
\]

Averaging over all \(i \in [n_h]\) gives
\[
\mathbb{E}\big[\bar{x}_{t+\frac{3}{4}}\big]
\;=\;
\mathbb{E}\big[\bar{x}_{t+\frac{2}{4}}\big].
\]

Now, we consider the \textbf{last step of aggregation} where the spokes aggregate models received from the hubs.

From the system dynamics, we have:
\[
X_{t+1} = W_{sh}\, X_{t+\frac{3}{4}},
\]
where \(W_{sh}\) is independent of \(X_{t+\frac{3}{4}}\). Taking expectations:
\[
\mathbb{E}\big[X_{t+1}\big] = \mathbb{E}\big[W_{sh}\big] \, \mathbb{E}\big[X_{t+\frac{3}{4}}\big].
\]

By the construction of \(W_{sh}\) as row-stochastic, we have:
\[
\sum_{j=1}^{n_h} W_{sh}^{(i,j)} = 1
\quad \text{for all spokes } i.
\]
Taking expectations and using symmetry (equal probability for all elements), let
\[
c = \mathbb{E}\big[W_{sh}^{(i,j)}\big].
\]
Then,
\[
\sum_{j=1}^{n_h} \mathbb{E}\big[W_{sh}^{(i,j)}\big]
= 1
\quad \Longrightarrow \quad
n_h\,c = 1
\quad \Longrightarrow \quad
c = \frac{1}{n_h}.
\]
Thus,
\[
\mathbb{E}[W_{sh}]
= \frac{1}{n_h}\,\mathbf{1}_{n_s}\,\mathbf{1}_{n_h}^T,
\]
where \(\mathbf{1}_{n_s}\) and \(\mathbf{1}_{n_h}\) are column vectors of ones of dimension \(n_s\) and \(n_h\), respectively. 

Substituting into the expectation, we get:
\[
\mathbb{E}\big[X_{t+1}\big]
= \frac{1}{n_h}\,\mathbf{1}_{n_s}\,\mathbf{1}_{n_h}^T\,
   \mathbf{1}_{n_h}\,\overline{x}_{t+\frac{3}{4}}
= \mathbf{1}_{n_s}\,\overline{x}_{t+\frac{3}{4}},
\]
since \(\mathbf{1}_{n_h}^T\,\mathbf{1}_{n_h} = n_h\).

Therefore,
\[
\mathbb{E}\big[\overline{x}_{t+1}\big] = \mathbb{E}\big[\overline{x}_{t+\frac{3}{4}}\big].
\]
Thus, we conclude that 
\[
\mathbb{E}\big[\overline{x}_{t+1}\big] = \mathbb{E}\big[\overline{x}_{t+\frac{3}{4}}\big] = \mathbb{E}\big[\overline{x}_{t+\frac{2}{4}}\big] = \mathbb{E}\big[\overline{x}_{t+\frac{1}{4}}\big].
\]

\end{proof}

\subsection{Proof of~\cref{lemma: 1b}} 
\paragraph{Stage One mixing: SPoke-to-Hub Push}
The models at the spokes after local training are denoted by \(x_{t+\tfrac{1}{4}}\). Each hub randomly samples \(b_{hs}\) spokes, and the model transfer from spoke \(j\) to hub \(i\) is represented by the indicator function \(I_j^i\). Hub \(i\) aggregates the \(b_{hs}\) collected models to produce \(x_{t+\tfrac{2}{4}}\). This proof bounds the consensus distance after mixing to that before it, that is: 
\begin{align*}
\frac{1}{n_h^2} \sum_{\substack{i, j \in [n_h] \\ i \neq j}} \mathbb{E}\!\left[\left\|x^{(i)}_{t+\tfrac{2}{4}} - x^{(j)}_{t+\tfrac{2}{4}}\right\|^2\right]
\;\le\;
\frac{\beta_{hs}}{n_s^2} \sum_{\substack{i, j \in [n_s] \\ i \neq j}} \mathbb{E}\!\left[\left\|x^{(i)}_{t+\tfrac{1}{4}} - x^{(j)}_{t+\tfrac{1}{4}}\right\|^2\right].
\end{align*}
where
\[
\beta_{hs} = \frac{1}{b_{hs}} \left[ 1 - \frac{b_{hs} - 1}{n_s - 1} \right].
\]
\begin{proof}
\begin{align*}
\frac{1}{n_h}\sum_{i \in [n_h]} \mathbb{E}\!\left[\left\|x^{(i)}_{t+\tfrac{2}{4}} - \bar{x}_{t+\tfrac{1}{4}}\right\|^2\right]
&= \frac{1}{n_h}\sum_{i} \mathbb{E}\!\left[\left\|\tfrac{1}{b_{hs}}\sum_{j \in [n_s]} \mathcal{I}_j^{(i)}\,x^{(j)}_{t+\tfrac{1}{4}} - \bar{x}_{t+\tfrac{1}{4}}\right\|^2\right]\\
&= \frac{1}{n_h\,b_{hs}^2}\sum_{i} \mathbb{E}\!\left[\left\|\sum_{j}\left(\mathcal{I}_j^{(i)}\,x^{(j)}_{t+\tfrac{1}{4}} - \bar{x}_{t+\tfrac{1}{4}}\right)\right\|^2\right]\\
&= \frac{1}{n_h\,b_{hs}^2}\sum_{i} \mathbb{E}\!\left[\sum_{j} \mathcal{I}_j^{(i)} \,\left\|x^{(j)}_{t+\tfrac{1}{4}} - \bar{x}_{t+\tfrac{1}{4}}\right\|^2\right] \\
&\quad+\; \frac{1}{n_h\,b_{hs}^2}\sum_{i} \mathbb{E}\!\Biggl[\sum_{j}\sum_{k \neq j} \mathcal{I}_j^{(i)}\,\mathcal{I}_k^{(i)}\,\Bigl\langle x^{(j)}_{t+\tfrac{1}{4}} - \bar{x}_{t+\tfrac{1}{4}}, \quad x^{(k)}_{t+\tfrac{1}{4}} - \bar{x}_{t+\tfrac{1}{4}} \Bigr\rangle\Biggr]\\
&= \frac{1}{n_h\,b_{hs}}\,\frac{1}{n_s}\sum_{i} \mathbb{E}\!\left[\sum_{j}\left\|x^{(j)}_{t+\tfrac{1}{4}} - \bar{x}_{t+\tfrac{1}{4}}\right\|^2\right] +\; \frac{1}{n_h\,b_{hs}}\,\frac{b_{hs}-1}{n_s\,(n_s-1)}(-1)
     \sum_{i} \mathbb{E}\!\left[\sum_{j}\left\|x^{(j)}_{t+\tfrac{1}{4}} - \bar{x}_{t+\tfrac{1}{4}}\right\|^2\right]
\end{align*}

where we utilize the fact that $\mathbb{E}[\mathcal{I}_j^{(i)}] =  \frac{b_{hs}}{n_s}$ and $\mathbb{E}[\mathcal{I}_j^{(i)} \mathcal{I}_k^{(i)}] =  \frac{b_{hs}}{n_s} \frac{b_{hs}-1}{n_s-1}$,  

Observe that
\[
\mathbb{E}\!\left[\sum_{j}\left\|x^{(j)}_{t+\tfrac{1}{4}} - \bar{x}_{t+\tfrac{1}{4}}\right\|^2\right]
\]  
is independent of $i$, therefore summing over all $i \in \left[n_h\right]$ scales the entire expression by a factor of $n_h$.  
Thus,

\begin{align}
\frac{1}{n_h}\sum_{j} \mathbb{E}\!\left[\left\|x^{(j)}_{t+\tfrac{2}{4}} - \bar{x}_{t+\tfrac{1}{4}}\right\|^2\right]
&= \frac{1}{n_s\,b_{hs}}\left(1 - \tfrac{b_{hs}-1}{n_s-1}\right)
   \mathbb{E}\!\left[\sum_{j}\left\|x^{(j)}_{t+\tfrac{1}{4}} - \bar{x}_{t+\tfrac{1}{4}}\right\|^2\right] \nonumber \\
&= \frac{\beta_{hs}}{n_s}\mathbb{E}\!\left[\sum_{j}\left\|x^{(j)}_{t+\tfrac{1}{4}} - \bar{x}_{t+\tfrac{1}{4}}\right\|^2\right].
\label{eqn: 1bi}
\end{align}
where 
\[
\beta_{hs} = \frac{1}{b_{hs}}\left(1 - \frac{b_{hs}-1}{n_s-1}\right)
\]

Noting that as \(\bar{y}\) is the minimizer of \(g(z) := \frac{1}{n} \sum_{i \in [n]} \mathbb{E} \left[\left\|y^{(i)} - z\right\|^2\right]\), we have
\begin{align}
\frac{1}{n} \sum_{i \in [n]} \mathbb{E} \left[\left\|y^{(i)} - \bar{y}\right\|^2\right] 
\leq \frac{1}{n} \sum_{i \in [n]} \mathbb{E} \left[\left\|y^{(i)} - \bar{x}\right\|^2\right].
\label{eqn: minimizer}
\end{align}
Substituting \(x\) to \(x_{t+\tfrac{1}{4}}\) and \(y\) to \(x_{t+\tfrac{2}{4}}\), and \(n\) to \(n_h\), and using (\ref{eqn: 1bi}), we obtain 

\begin{align}
\frac{1}{n_h}\sum_{i=1}^{n_h} \mathbb{E}\!\left[\left\|x^{(i)}_{t+\tfrac{2}{4}} - \bar{x}_{t+\tfrac{2}{4}}\right\|^2\right]
\;\le\;
\frac{1}{n_h}\sum_{i=1}^{n_h} \mathbb{E}\!\left[\left\|x^{(i)}_{t+\tfrac{2}{4}} - \bar{x}_{t+\tfrac{1}{4}}\right\|^2\right]
\;=\;
\frac{\beta_{hs}}{n_s}\sum_{i=1}^{n_s} \mathbb{E}\!\left[\left\|x^{(i)}_{t+\tfrac{1}{4}} - \bar{x}_{t+\tfrac{1}{4}}\right\|^2\right].
\label{eqn: 1b-i-inequality}
\end{align}

Using~\cref{lemma-5}, we obtain,
\begin{align}
\frac{1}{n_h^2} \sum_{\substack{i, j \in [n_h] \\ i \neq j}} \mathbb{E}\!\left[\left\|x^{(i)}_{t+\tfrac{2}{4}} - x^{(j)}_{t+\tfrac{2}{4}}\right\|^2\right]
\;\le\;
\frac{\beta_{hs}}{n_s^2} \sum_{\substack{i, j \in [n_s] \\ i \neq j}} \mathbb{E}\!\left[\left\|x^{(i)}_{t+\tfrac{1}{4}} - x^{(j)}_{t+\tfrac{1}{4}}\right\|^2\right].
\label{eqn-1b-final}
\end{align}
\end{proof}
\paragraph{Stage Two mixing: Hub Gossip}
During the hub-gossip stage, every hub shares its models with \(b_{hh}\) other hubs, all having a constant outdegree. However, the indegree of the hubs is a variable \(A^{(i)}\). This is exactly how nodes communicate in \ellocal. Then the consensus distance among the hub models after then gossip stage is bound by:
\begin{align*}
    \frac{1}{n_h^2} \sum_{\substack{i, j \\ i \neq j}} \mathbb{E} \left[ \left\| x_{t+\tfrac{3}{4}}^{(i)} - x_{t+\tfrac{3}{4}}^{(j)} \right\|^2 \right] 
    &\leq \beta_{hh} \cdot \frac{1}{n_h^2} \sum_{\substack{i, j \\ i \neq j}} \mathbb{E} \left[ \left\| x_{t+\tfrac{2}{4}}^{(i)} - x_{t+\tfrac{2}{4}}^{(j)} \right\|^2 \right],
\end{align*}
where
\[
\beta_{hh} = \frac{1}{b_{hh}} \left( 1 - \left( 1 - \frac{b_{hh}}{n_h - 1} \right)^{n_h} \right) - \frac{1}{n_h-1}.
\]

\begin{proof}
\begin{align*}
&\frac{1}{n_h}\sum_{i\in [n_h]}\mathbb{E}\left[\left\|x_{t+\frac{3}{4}}^{(i)}-\bar{x}_{t+\frac{2}{4}}\right\|^2\right] = \frac{1}{n_h}\sum_{i\in [n_h]}\mathbb{E}\left[\left\|\frac{1}{A^{(i)}+1}\left(x_{t+\frac{2}{4}}^{(i)}+\sum_{j\in [n_h] \backslash \{i\}}\mathcal{I}_j^{(i)}x_{t+\frac{2}{4}}^{(j)}\right)-\bar{x}_{t+\frac{2}{4}}\right\|^2\right]\\
&= \frac{1}{n_h}\sum_{i\in [n_h]}\mathbb{E}\left[\mathbb{E}\left[\left\|\frac{1}{A^{(i)}+1}\left(x_{t+\frac{2}{4}}^{(i)}+\sum_{j}\mathcal{I}_j^{(i)} x_{t+\frac{2}{4}}^{(j)}\right)-\bar{x}_{t+\frac{2}{4}} \right\|^2|A^{(i)}\right]\right]\\
&= \frac{1}{n_h}\sum_{i\in [n_h]}\mathbb{E}\left[\mathbb{E}\left[\left\|\frac{1}{A^{(i)}+1}\left((x_{t+\frac{2}{4}}^{(i)} - x_{t+\frac{2}{4}}^{(i)} )+\sum_{j}\mathcal{I}_j (x_{\frac{2}{4}}^{(j)}-\bar{x}_{t+\frac{2}{4}})\right)\right\|^2 |A^{(i)}\right]\right]\\
&= \frac{1}{n_h}\sum_{i\in [n_h]}\mathbb{E}\left[\frac{1}{A^{(i)}+1}\mathbb{E}\left[\left\|x_{t+\frac{2}{4}}^{(i)} - x_{t+\frac{2}{4}}^{(i)} \right\|^2+\sum_{j}\mathcal{I}_j \left\| (x_{\frac{2}{4}}^{(j)}-\bar{x}_{t+\frac{2}{4}})\right\|^2 \right]|A^{(i)}\right]\\
&=\frac{1}{n_h}\sum_{i\in[n_h]}\mathbb{E}\left[\frac{1}{(A^{(i)}+1)^{2}}\mathbb{E}\left[2\sum_{j\ne i}\mathcal{I}_{j}^{(i)}\langle x_{\frac{2}{4}}^{(i)}-\bar{x}_{t+\frac{2}{4}},x_{t+\frac{2}{4}}^{(j)}-\bar{x}_{t+\frac{2}{4}}\rangle+\sum_{j\ne i}\sum_{k\ne i,k\ne j}\mathcal{I}_{j}^{(i)}\mathcal{I}_{k}^{(i)}\langle x_{\frac{2}{4}}^{(j)}-\bar{x}_{t+\frac{2}{4}},x_{t+\frac{2}{4}}^{(k)}-\bar{x}_{t+\frac{2}{4}}\rangle |A^{(i)}\right]\right]
\end{align*}

Taking the expectation inside, we obtain
\begin{align*}
\frac{1}{n_h}\sum_{i\in[n_h]}\mathbb{E}[\left|x_{t+\frac{3}{4}}^{(i)}-\bar{x}_{t+\frac{2}{4}}\right|^{2}] &= \frac{1}{n_h}\sum_{i\in[n_h]}\mathbb{E}\left[\frac{1}{(A^{(i)}+1)^{2}}\left(\left\|x_{t+\frac{2}{4}}^{(i)}-\bar{x}_{t+\frac{2}{4}}\right\|^{2}+\sum_{j\ne i}\mathbb{E}[\mathcal{I}_{j}^{(i)}|A^{(i)}]\left\|x_{t+\frac{2}{4}}^{(j)}-\bar{x}_{t+\frac{2}{4}}\right\|^{2}\right)\right] \\
&+ \frac{1}{n_h}\sum_{i\in[n_h]}\mathbb{E}\left[\frac{1}{(A^{(i)}+1)^{2}}\left(2\sum_{j\ne i}\mathbb{E}[\mathcal{I}_{j}^{(i)}|A^{(i)}]\langle x_{t+\frac{2}{4}}^{(i)}-\bar{x}_{t+\frac{2}{4}},x_{t+\frac{2}{4}}^{(j)}-\bar{x}_{t+\frac{2}{4}}\rangle\right)\right] \\
&+ \frac{1}{n_h}\sum_{i\in[n_h]}\mathbb{E}\left[\frac{1}{(A^{(i)}+1)^{2}}\left(\sum_{j\ne i,k\ne i,k\ne j}\mathbb{E}[\mathcal{I}_{j}^{(i)}\mathcal{I}_{k}^{(i)}|A^{(i)}]\langle x_{t+\frac{2}{4}}^{(j)}-\bar{x}_{t+\frac{2}{4}},x_{t+\frac{2}{4}}^{(k)}-\bar{x}_{t+\frac{2}{4}}\rangle\right)\right].
\end{align*}

Observe that $\mathbb{E}[\mathcal{I}_{j}^{(i)}|A^{(i)}]$ represents the probability of node $j$ selecting node $i$, given that a total of $A^{(i)}$ nodes select $i$. Thus,
\begin{align*}
&\mathbb{E}[\mathcal{I}_{j}^{(i)}|A^{(i)}] = \frac{A^{(i)}}{n_h-1}
\end{align*}
Similarly, $\mathcal{I}_{j}^{(i)}\mathcal{I}_{k}^{(i)}$ equals 1 only when both $j$ and $k$ choose $i$, hence
\begin{align*}
&\mathbb{E}\left[\mathcal{I}_{j}^{(i)}\mathcal{I}_{k}^{(i)}|A^{(i)}\right] = \frac{A^{(i)}(A^{(i)}-1)}{(n_h-1)(n_h-2)}. 
\end{align*}
Also, note that
\begin{align*}
&\sum_{j\neq i}\langle x_{t+\frac{2}{4}}^{(i)}-\bar{x}_{t+\frac{2}{4}},x_{t+\frac{2}{4}}^{(j)}-\bar{x}_{t+\frac{2}{4}}\rangle = \langle x_{t+\frac{2}{4}}^{(i)}-\bar{x}_{t+\frac{2}{4}},\sum_{j\neq i}(x_{t+\frac{2}{4}}^{(j)}-\bar{x}_{t+\frac{2}{4}}\rangle = -\left\|x_{t+\frac{2}{4}}^{(i)}-\bar{x}_{t+\frac{2}{4}}\right\|^2 \\
\end{align*}
and
\begin{align*}
&\sum_{j\neq i}\sum_{k\neq i,k\neq j}\langle x_{t+\frac{2}{4}}^{(j)}-\bar{x}_{t+\frac{2}{4}},x_{t+\frac{2}{4}}^{(k)}-\bar{x}_{t+\frac{2}{4}}\rangle = \sum_{j\neq i}\left\langle x_{t+\frac{2}{4}}^{(j)}-\bar{x}_{t+\frac{2}{4}},\sum_{k\neq i,k\neq j}(x_{t+\frac{2}{4}}^{(k)}-\bar{x}_{t+\frac{2}{4}})\right\rangle \\
&= \sum_{j\neq i}\left\langle x_{t+\frac{2}{4}}^{(j)}-\bar{x}_{t+\frac{2}{4}},(x_{t+\frac{2}{4}}^{(i)}-\bar{x}_{t+\frac{2}{4}})+(x_{t+\frac{2}{4}}^{(j)}-\bar{x}_{t+\frac{2}{4}})\right\rangle \\
&= \|x_{t+\frac{2}{4}}^{(i)}-\bar{x}_{t+\frac{2}{4}}\|^2-\sum_{j\neq i}\|x_{t+\frac{2}{4}}^{(j)}-\bar{x}_{t+\frac{2}{4}}\|^2
\end{align*}

Bringing everything together, we obtain
\begin{align*}
\frac{1}{n_h}\sum_{i\in[n_h]}\mathbb{E}\left[\left\|x_{t+\frac{3}{4}}^{(i)}-\bar{x}_{t+\frac{2}{4}}\right\|^{2}\right] &= \frac{1}{n_h}\sum_{i\in[n_h]}\mathbb{E}\left[\frac{1}{(A^{(i)}+1)^{2}}\left(\left\|x_{t+\frac{2}{4}}^{(i)}-\bar{x}_{t+\frac{2}{4}}\right\|^{2}+\frac{A^{(i)}}{n_h-1}\sum_{j\ne i}\left\|x_{t+\frac{2}{4}}^{(j)}-\bar{x}_{t+\frac{2}{4}}\right\|^{2}\right)\right] \\
&+ \frac{1}{n_h}\sum_{i\in[n_h]}\mathbb{E}\left[\frac{1}{(A^{(i)}+1)^{2}}\left(-\frac{2A^{(i)}}{n_h-1}\left\|x_{t+\frac{2}{4}}^{(i)}-\bar{x}_{t+\frac{2}{4}}\right\|^{2}\right)\right] \\
&+ \frac{1}{n_h}\sum_{i\in[n_h]}\mathbb{E}\left[\frac{1}{(A^{(i)}+1)^{2}}\left(\frac{A^{(i)}(A^{(i)}-1)}{(n_h-1)(n_h-2)}\left(\left\|x_{t+\frac{2}{4}}^{(i)}-\bar{x}_{t+\frac{2}{4}}\right\|^{2}-\sum_{j\ne i}\left\|x_{t+\frac{2}{4}}^{(j)}-\bar{x}_{t+\frac{2}{4}}\right\|^{2}\right)\right)\right] \\
&= \frac{1}{n_h}\sum_{i\in[n_h]}\left\|x_{t+\frac{2}{4}}^{(i)}-\bar{x}_{t+\frac{2}{4}}\right\|^{2}\mathbb{E}\left[\frac{1}{(A^{(i)}+1)^{2}}\left(1-\frac{2A^{(i)}}{n_h-1}+\frac{A^{(i)}(A^{(i)}-1)}{(n_h-1)(n_h-2)}\right)\right] \\
&+ \frac{1}{n_h}\sum_{i\in[n_h]}\mathbb{E}\left[\frac{1}{(A^{(i)}+1)^{2}}\left(\frac{A^{(i)}}{n_h-1}-\frac{A^{(i)}(A^{(i)}-1)}{(n_h-1)(n_h-2)}\right)\right]\sum_{j\ne i}\left\|x_{t+\frac{2}{4}}^{(j)}-\bar{x}_{t+\frac{2}{4}}\right\|^{2}
\end{align*}

Observe that, due to symmetry, the distribution of $A^{(i)}$ is identical to that of $A^{(j)}$ for any $i, j \in [n_h]$. Hence,

\begin{align*}
&\frac{1}{n_h}\sum_{i\in[n_h]}\mathbb{E}\left[\left\|x_{t+\frac{3}{4}}^{(i)}-\bar{x}_{t+\frac{2}{4}}\right\|^{2}\right] \\
&= \frac{1}{n_h}\sum_{i\in[n_h]}\left\|x_{t+\frac{2}{4}}^{(i)}-\bar{x}_{t+\frac{2}{4}}\right\|^{2}\mathbb{E}\left[\frac{1}{(A^{(1)}+1)^{2}}\left(1-\frac{2A^{(1)}}{n_h-1}+\frac{A^{(1)}(A^{(1)}-1)}{(n_h-1)(n_h-2)}+\frac{A^{(1)}}{n_h-1}-\frac{A^{(1)}(A^{(1)}-1)}{(n_h-2)}\right)\right] 
\end{align*}

Now note that

\begin{align*}
    1 - \frac{2A^{(1)}}{n_h-1} + \frac{A^{(1)}(A^{(1)}-1)}{(n_h-1)(n_h-2)} + A^{(1)} - \frac{A^{(1)}(A^{(1)}-1)}{n_h-2} = 1 + A^{(1)} - \frac{A^{(1)^2} + A^{(1)}}{n_h-1} = (1 + A^{(1)})\left(1 - \frac{A^{(1)}}{n_h-1}\right)
\end{align*}

Thus

\begin{align*}
    \frac{1}{n_h}\sum_{i\in[n_h]}\mathbb{E}\left[\left\|x_{t+\frac{3}{4}}^{(i)}-\bar{x}_{t+\frac{2}{4}}\right\|^{2}\right] = \frac{1}{n_h}\sum_{i\in[n_h]}||x_{t+\frac{2}{4}}^{(i)}-\bar{x}_{t+\frac{2}{4}}||^{2}\left[\mathbb{E}\left[\frac{1}{A^{(1)}+1}\right]-\frac{1}{n_h-1}\cdot\mathbb{E}\left[\frac{A^{(1)}}{A^{(1)}+1}\right]\right]
\end{align*}

Observe that since each node $j \neq 1$ independently and uniformly selects a set of $b_{hh}$ nodes, $A^{(1)}$ follows a binomial distribution with parameters $n_h - 1$ and $\frac{b_{hh}}{n_h - 1}$. Thus, for $b_{hh} > 0$, we have

\begin{align*}
\mathbb{E}\left[\frac{1}{A^{(1)}+1}\right] &= \sum_{k=0}^{n_h-1}\frac{1}{k+1}\binom{n_h-1}{k}\left(\frac{b_{hh}}{n_h-1}\right)^{k}\left(1-\frac{b_{hh}}{n_h-1}\right)^{n_h-1-k} \\
&= \frac{n_h-1}{b_{hh} n_h}\sum_{k=0}^{n_h-1}\binom{n_h}{k+1}\left(\frac{b_{hh}}{n_h-1}\right)^{k+1}\left(1-\frac{b_{hh}}{n_h-1}\right)^{n_h-1-k} \\
&= \frac{n_h-1}{b_{hh}n_h}\left(1-\left(1-\frac{b_{hh}}{n_h-1}\right)^{n_h}\right)
\end{align*}

Also noting that

\begin{align*}
    \mathbb{E}\left[\frac{A^{(1)}}{A^{(1)}+1}\right] = 1 - \mathbb{E}\left[\frac{1}{A^{(1)}+1}\right],
\end{align*}

we obtain

\begin{align}
    \frac{1}{n_h}\sum_{i\in[n_h]}\mathbb{E}\left[\left\|x_{t+\frac{3}{4}}^{(i)}-\bar{x}_{t+\frac{2}{4}}\right\|^{2}\right] = \left[\frac{1}{b_{hh}}\left(1-\left(1-\frac{b_{hh}}{n_h-1}\right)^{n_h}\right)-\frac{1}{n_h-1}\right]\frac{1}{n_h}\sum_{i\in[n_h]}\left\|x_{t+\frac{2}{4}}^{(i)}-\bar{x}_{t+\frac{2}{4}}\right\|^{2}
\label{eqn:1b-ii-equal}
\end{align}

\text{we obtain that}

Noting that as \(\bar{y}\) is the minimizer of \(g(z) := \frac{1}{n} \sum_{i \in [n]} \mathbb{E} \left[\left\|y^{(i)} - z\right\|^2\right]\), following the logic in (\ref{eqn: minimizer}), and using~\cref{lemma-5}, we convert the equality in (\ref{eqn:1b-ii-equal}) to the following inequality:

\begin{align}
\frac{1}{n_h^2}\sum_{i,j\in[n_h]}\mathbb{E}\left[\left\|x_{t+\frac{3}{4}}^{(i)}-x_{t+\frac{3}{4}}^{(j)}\right\|^{2}\right] \leq \left[\frac{1}{b_{hh}}\left(1-\left(1-\frac{b_{hh}}{n_h-1}\right)^{n_h}\right)-\frac{1}{n_h-1}\right]\frac{1}{n_h^2}\sum_{i,j\in[n_h]}\mathbb{E}\left[\left\|x_{t+\frac{2}{4}}^{(i)}-x_{t+\frac{2}{4}}^{(j)}\right\|^{2}\right] 
\label{eqn: weired-glitch}
\end{align}

which is the desired result.
\end{proof}

\paragraph{Stage Three Mixing: Hub-to-Spoke Pull}
The models at the hubs after hub gossip are represented as \( x_{t+\tfrac{3}{4}} \). Each spoke independently selects \( b_{sh} \) hubs at random, with the model transfer from hub \( j \) to spoke \( i \) indicated by the function \(\mathcal{I}_j^i\). Spoke \( i \) then aggregates the \( b_{sh} \) received models to obtain \( x_{t+1} \). This proof establishes an upper bound on the consensus distance among the spoke models after the final aggregation stage relative to its value before aggregation, that is,
\begin{align*}
\frac{1}{n_s^2} \sum_{\substack{i, j \in [n_s] \\ i \neq j}} \mathbb{E}\!\left[\left\|x^{(i)}_{t+1} - x^{(j)}_{t+1}\right\|^2\right]
\;\le\;
\frac{\beta_{sh}}{n_h^2} \sum_{\substack{i, j \in [n_h] \\ i \neq j}} \mathbb{E}\!\left[\left\|x^{(i)}_{t+\tfrac{3}{4}} - x^{(j)}_{t+\tfrac{3}{4}}\right\|^2\right].
\end{align*}
where
\[
\beta_{sh} = \frac{1}{b_{sh}} \left[ 1 - \frac{b_{sh} - 1}{n_h - 1} \right].
\]

\begin{proof}
\begin{align*}
\frac{1}{n_s}\sum_{i \in [n_s]} \mathbb{E}\!\left[\left\|x^{(i)}_{t+1} - \bar{x}_{t+\tfrac{3}{4}}\right\|^2\right]
&= \frac{1}{n_s}\sum_{i} \mathbb{E}\!\left[\left\|\tfrac{1}{b_{sh}}\sum_{j} \mathcal{I}_j^{(i)}\,x^{(j)}_{t+\tfrac{3}{4}} - \bar{x}_{t+\tfrac{3}{4}}\right\|^2\right]\\
&= \frac{1}{n_s\,b_{sh}^2}\sum_{i} \mathbb{E}\!\left[\left\|\sum_{j}\left(\mathcal{I}_j^{(i)}\,x^{(j)}_{t+\tfrac{3}{4}} - \bar{x}_{t+\tfrac{3}{4}}\right)\right\|^2\right]\\
&= \frac{1}{n_s\,b_{sh}^2}\sum_{i} \mathbb{E}\!\left[\sum_{j} \mathcal{I}_j^{(i)} \,\left\|x^{(j)}_{t+\tfrac{3}{4}} - \bar{x}_{t+\tfrac{3}{4}}\right\|^2\right] \\
&\quad+\; \frac{1}{n_s\,b_{sh}^2}\sum_{i} \mathbb{E}\!\Biggl[\sum_{j}\sum_{k \neq j} \mathcal{I}_j^{(i)}\,\mathcal{I}_k^{(i)}\,\Bigl\langle x^{(j)}_{t+\tfrac{3}{4}} - \bar{x}_{t+\tfrac{3}{4}}, \quad x^{(k)}_{t+\tfrac{3}{4}} - \bar{x}_{t+\tfrac{3}{4}} \Bigr\rangle\Biggr]\\
&= \frac{1}{n_s\,b_{sh}}\,\frac{1}{n_h}\sum_{i} \mathbb{E}\!\left[\sum_{j}\left\|x^{(j)}_{t+\tfrac{3}{4}} - \bar{x}_{t+\tfrac{3}{4}}\right\|^2\right] +\; \frac{1}{n_s\,b_{sh}}\,\frac{b_{sh}-1}{n_h\,(n_h-1)}(-1)
     \sum_{i} \mathbb{E}\!\left[\sum_{j}\left\|x^{(j)}_{t+\tfrac{3}{4}} - \bar{x}_{t+\tfrac{3}{4}}\right\|^2\right]\\
\end{align*}
where we utilize the fact that $\mathbb{E}[\mathcal{I}_j^{(i)}] =  \frac{b_{sh}}{n_h}$ and $\mathbb{E}[\mathcal{I}_j^{(i)} \mathcal{I}_k^{(i)}] =  \frac{b_{sh}}{n_h} \frac{b_{sh}-1}{n_h-1}$,  

Just like for stage 1 aggregation, observe that
\[
\mathbb{E}\!\left[\sum_{j}\left\|x^{(j)}_{t+\tfrac{3}{4}} - \bar{x}_{t+\tfrac{3}{4}}\right\|^2\right]
\]  
is independent of $i$, therefore summing over all $i \in \left[n_s\right]$ scales the entire expression by a factor of $n_s$.  
Thus,  
\begin{align}
\frac{1}{n_s}\sum_{i \in [n_s]} \mathbb{E}\!\left[\left\|x^{(i)}_{t+1} - \bar{x}_{t+\tfrac{3}{4}}\right\|^2\right]     
&= \frac{1}{n_h\,b_{sh}}\left(1 - \tfrac{b_{sh}-1}{n_h-1}\right)
   \mathbb{E}\!\left[\sum_{j}\left\|x^{(j)}_{t+\tfrac{3}{4}} - \bar{x}_{t+\tfrac{3}{4}}\right\|^2\right] \nonumber \\
&= \frac{\beta_{hs}}{n_h} \mathbb{E}\!\left[\sum_{j}\left\|x^{(j)}_{t+\tfrac{3}{4}} - \bar{x}_{t+\tfrac{3}{4}}\right\|^2\right].
\label{eqn: 1b=iii}
\end{align}

where \(\beta_{sh}\):
\[
\beta_{sh} = \frac{1}{b_{sh}}\left(1 - \frac{b_{sh}-1}{n_h-1}\right).
\]

Using the minimizer property described in (\ref{eqn: minimizer}), and applying~\cref{lemma-5}, we finally obtain:
\begin{align}
\frac{1}{n_s^2} \sum_{\substack{i, j \in [n_s] \\ i \neq j}} \mathbb{E}\!\left[\left\|x^{(i)}_{t+1} - x^{(j)}_{t+1}\right\|^2\right]
\;\le\;
\frac{\beta_{sh}}{n_h^2} \sum_{\substack{i, j \in [n_h] \\ i \neq j}} \mathbb{E}\!\left[\left\|x^{(i)}_{t+\tfrac{3}{4}} - x^{(j)}_{t+\tfrac{3}{4}}\right\|^2\right].
\label{eqn: 1b-iii-in-2}
\end{align}

\end{proof}

\subsection{Proof of~\cref{lemma: 1c}}
\paragraph{Stage 1 Spoke-to-Hub Push}
We have: 
\begin{align*}
    \mathbb{E} \left[ \left\| \bar{x}_{t+\frac{2}{4}} - \bar{x}_{t+\frac{1}{4}} \right\|^2 \right] 
    =\frac{\beta_{hs}}{n_s n_h}\sum_{i} \mathbb{E} \left[ \left\| x_{t+\frac{1}{4}}^{(i)} - \bar{x}_{t+\frac{1}{4}} \right\|^2 \right]
\end{align*}
\begin{proof}
Note that we can expand the norm as follows:
\begin{align}
    \mathbb{E} \left[ \left\| \bar{y} - \bar{x} \right\|^2 \right] &= \mathbb{E} \left[ \left\| \frac{1}{n} \sum_{i} y^{(i)} - \bar{x} \right\|^2 \right]\nonumber \\ 
    &= \frac{1}{n^2} \sum_{i} \mathbb{E} \left[ \left\| y^{(i)} - \bar{x} \right\|^2 \right] + \frac{1}{n^2} \sum_{i \neq j} \mathbb{E} \left[ \left< y^{(i)} - \bar{x}, y^{(j)} - \bar{x} \right> \right] 
\label{eqn: lemma-1c-begin}
\end{align}
For the first stage of communication from spokes to hubs, we denote ${x}_{t+\frac{2}{4}}$ as $y$ and ${x}_{t+\frac{1}{4}}$ as $x$, replacing $n$ with $n_h$. For the $i$-th hub, we have:
\begin{align*}
    x_{t+\frac{2}{4}}^{(i)} - \bar{x}_{t+\frac{1}{4}} = \frac{1}{b_{hs}} \sum_{k} \mathcal{I}_{k}^{(i)} (x_{t+\frac{1}{4}}^{(k)} - \bar{x}_{t+\frac{1}{4}})
\end{align*}
where $\mathcal{I}_{k}^{(i)}$ is an indicator function that represents the connectivity between hub $i$ and spoke $k$.
We can thus write the second term in (\ref{eqn: lemma-1c-begin}) as

\begin{align}
\frac{1}{n^2}\sum_{i \neq j} \mathbb{E}\left[\langle y^{(i)} - \bar{x}, y^{(j)} - \bar{x} \rangle\right] &= \frac{1}{n_h^2}\sum_{\substack{i \in [n_h] \\i \neq j}} \mathbb{E}\left[\langle x_{t+\frac{2}{4}}^{(i)} - \bar{x}_{t+\frac{1}{4}}, x_{t+\frac{2}{4}}^{(j)} - \bar{x}_{t+\frac{1}{4}} \rangle\right] \nonumber \\
&= 
\frac{2}{n_h^2} \sum_{i \ne j} \sum_{k \in [n_s]} \sum_{l \in [n_s]} \mathbb{E} \left[ \frac{\mathcal{I}_{k}^{(i)} \mathcal{I}_{l}^{(j)}}{b_{hs}^2} \langle x_{t+\frac{1}{4}}^{(k)} - \bar{x}_{t+\frac{1}{4}}, x_{t+\frac{1}{4}}^{(l)} - \bar{x}_{t+\frac{1}{4}} \rangle \right]
\label{eqn: lemma-1c-i-kl}
\end{align}

Now note that by symmetry, for any $i, j \in [n_h]$, we have,

\begin{align*}
    \mathbb{E}\left[\mathcal{I}_{k}^{(i)}\mathcal{I}_{l}^{(j)}\right] = \mathbb{E}\left[\mathcal{I}_{3}^{(1)}\mathcal{I}_{4}^{(2)}\right]
\end{align*}

This implies that all three terms in (\ref{eqn: lemma-1c-i-kl}) can be written as,

\begin{align*}
    &c \cdot \mathbb{E} \left[\sum_{k \in [n_s]} \sum_{l \in [n_s]} \left\langle x_{t+\frac{1}{4}}^{(k)} - \bar{x}_{t+\frac{1}{4}}, x_{t+\frac{1}{4}}^{(l)} - \bar{x}_{t+\frac{1}{4}} \right\rangle \right]\\
    &=c \cdot \mathbb{E} \left[ \sum_l \left\| x_{t+\frac{1}{4}}^{(l)} - \bar{x}_{t+\frac{1}{4}} \right\|^2 + \sum_{k \ne l} \left\langle x_{t+\frac{1}{4}}^{(k)} - \bar{x}_{t+\frac{1}{4}}, x_{t+\frac{1}{4}}^{(l)} - \bar{x}_{t+\frac{1}{4}} \right\rangle \right] \\
    &= c \cdot \mathbb{E} \left[ \sum_l \left\| x_{t+\frac{1}{4}}^{(l)} - \bar{x}_{t+\frac{1}{4}} \right\|^2 - \sum_l \left\| x_{t+\frac{1}{4}}^{(l)} - \bar{x}_{t+\frac{1}{4}} \right\|^2\right] \\
    &= 0
\end{align*}

Therefore, from equation (\ref{eqn: lemma-1c-begin}), we obtain
\begin{align}
    \mathbb{E} \left[ \left\| \bar{x}_{t+\frac{2}{4}} - \bar{x}_{t+\frac{1}{4}} \right\|^2 \right] &= \frac{1}{n_h^2} \sum_{i} \mathbb{E} \left[ \left\| x_{t+\frac{2}{4}}^{(i)} - \bar{x}_{t+\frac{1}{4}} \right\|^2 \right] \nonumber \\
    &= \frac{\beta_{hs}}{n_s n_h}\sum_{i} \mathbb{E} \left[ \left\| x_{t+\frac{1}{4}}^{(i)} - \bar{x}_{t+\frac{1}{4}} \right\|^2 \right]
\end{align}
where we make use of (\ref{eqn: 1bi}).

\end{proof}

\paragraph{Stage 2 mixing: Hub Gossip}
Here we prove that
\begin{align*}
\mathbb{E}\left[\left\|x_{t+\frac{3}{4}} - \bar{x}_{t+\frac{2}{4}}\right\|^2\right] 
& \leq \frac{\beta_{hh}}{n_h^2} \sum_{i \in [n_h]} \left\|x^{(i)}_{t+\frac{2}{4}} - \bar{x}_{t+\frac{2}{4}}\right\|^2
\end{align*}
\begin{proof}
For the second stage of aggregation, we have: 
\begin{align}
\mathbb{E}\left[\left\|x_{t+\frac{3}{4}} - \bar{x}_{t+\frac{2}{4}}\right\|^2\right]
&= \mathbb{E}\left[\left\|\frac{1}{n_h}\sum_{i \in [n_h]} \left(x_{t+\frac{3}{4}}^{(i)} - \bar{x}_{t+\frac{2}{4}}\right)\right\|^2\right] \nonumber \\
&= \frac{1}{n^2} \sum_{i \in [n_h]} \mathbb{E}\left[\left\|x_{t+\frac{3}{4}}^{(i)} - \bar{x}_{t+\frac{2}{4}}\right\|^2\right] + \frac{1}{n_h^2} \sum_{i \neq j} \mathbb{E}\left[\left\langle x_{t+\frac{3}{4}}^{(i)} - \bar{x}_{t+\frac{2}{4}}, x_{t+\frac{3}{4}}^{(j)} - \bar{x}_{t+\frac{2}{4}}\right\rangle\right]
\label{eqn:1-c-ii-begin}
\end{align}

Now recall that

\begin{align*}
    x_{t+\frac{3}{4}}^{(i)} - \bar{x}_{t+\frac{2}{4}} = \frac{1}{A^{(i)} + 1} \left( (x^{(i)}_{t+\frac{2}{4}} - \bar{x}_{t+\frac{2}{4}}) + \sum_{k \in [n_h] \setminus \{i\}} \mathcal{I}_{k}^{(i)} (x^{(k)}_{t+\frac{2}{4}} - \bar{x}_{t+\frac{2}{4}}) \right)
\end{align*}

This implies that

\begin{align}
\frac{1}{n_h^2} \sum_{i \neq j} \mathbb{E}\left[\left\langle x_{t+\frac{3}{4}}^{(i)} - \bar{x}_{t+\frac{2}{4}}, x_{t+\frac{3}{4}}^{(j)} - \bar{x}_{t+\frac{2}{4}}\right\rangle\right]
&= \frac{1}{n_{h}^2} \sum_{i \in [n_{h}]} \sum_{j \neq i} \mathbb{E}\left[\frac{1}{(A^{(i)} + 1)(A^{(j)} + 1)} \left\langle x^{(i)}_{t+\frac{2}{4}} - \bar{x}_{t+\frac{2}{4}}, x^{(j)}_{t+\frac{2}{4}} - \bar{x}_{t+\frac{2}{4}}\right\rangle\right] \nonumber\\
&+ \frac{2}{n_{h}^2} \sum_{i \in [n_{h}]} \sum_{j \neq i} \sum_{k \neq i,k \neq j} \mathbb{E}\left[\frac{\mathcal{I}_{k}^{(i)}}{(A^{(i)} + 1)(A^{(j)} + 1)} \left\langle x^{(k)}_{t+\frac{2}{4}} - \bar{x}_{t+\frac{2}{4}}, x^{(j)}_{t+\frac{2}{4}} - \bar{x}_{t+\frac{2}{4}}\right\rangle\right]\nonumber \\
&+ \frac{2}{n_{h}^2} \sum_{i \in [n_{h}]} \sum_{j \neq i} \sum_{k \neq i,k \neq j} \sum_{l \neq i,l \neq j,l \neq k} \mathbb{E}\left[\frac{\mathcal{I}_{k}^{(i)}\mathcal{I}_{l}^{(j)}}{(A^{(i)} + 1)(A^{(j)} + 1)} \left\langle x^{(k)}_{t+\frac{2}{4}} - \bar{x}_{t+\frac{2}{4}}, x^{(l)}_{t+\frac{2}{4}} - \bar{x}_{t+\frac{2}{4}}\right\rangle\right]
\label{eqn-1-c-ii}
\end{align}

Now note that by symmetry, for any  $i,j \in [n_h]$, we have

\begin{align*}
    \mathbb{E}\left[\frac{1}{(A^{(i)}+1)(A^{(j)}+1)}\right] = \mathbb{E}\left[\frac{1}{(A^{(1)}+1)(A^{(2)}+1)}\right]
\end{align*}
Similarly

\begin{align*}
    \mathbb{E}\left[\frac{\mathcal{I}_{k}^{(i)}}{(A^{(i)}+1)(A^{(j)}+1)}\right] = \mathbb{E}\left[\frac{\mathcal{I}_{3}^{(1)}}{(A^{(1)}+1)(A^{(2)}+1)}\right]
\end{align*}

and

\begin{align*}
    \mathbb{E}\left[\frac{\mathcal{I}_{k}^{(i)}\mathcal{I}_{l}^{(j)}}{(A^{(i)}+1)(A^{(j)}+1)}\right] = \mathbb{E}\left[\frac{\mathcal{I}_{3}^{(1)}\mathcal{I}_{4}^{(2)}}{(A^{(1)}+1)(A^{(2)}+1)}\right]
\end{align*}

This implies that all three terms in (\ref{eqn-1-c-ii}) can be written as

\begin{align*}
    c\sum_{i\in[n]}\sum_{j\ne i}\langle x^{(i)}_{t+\frac{2}{4}} - \bar{x}_{t+\frac{2}{4}},x^{(j)}_{t+\frac{2}{4}} - \bar{x}_{t+\frac{2}{4}}\rangle
\end{align*}

where c is a positive constant. We also have

\begin{align*}
\sum_{i\in[n_h]}\sum_{j\neq i} \langle x^{(i)}_{t+\frac{2}{4}} - \bar{x}_{t+\frac{2}{4}}, x^{(j)}_{t+\frac{2}{4}} - \bar{x}_{t+\frac{2}{4}} \rangle &= \sum_{i\in[n_h]} \left\langle x^{(i)}_{t+\frac{2}{4}} - \bar{x}_{t+\frac{2}{4}}, \sum_{j\neq i} (x^{(j)}_{t+\frac{2}{4}} - \bar{x}_{t+\frac{2}{4}}) \right\rangle \\
&= - \sum_{i\in[n_h]} \left\| x^{(i)}_{t+\frac{2}{4}} - \bar{x}_{t+\frac{2}{4}} \right\|^2.
\end{align*}

Therefore all the terms in (\ref{eqn-1-c-ii}) are non-positive. Combining this with (\ref{eqn:1-c-ii-begin}), we obtain that

\begin{align*}
\mathbb{E}\left[\left\|x_{t+\frac{3}{4}} - \bar{x}_{t+\frac{2}{4}}\right\|^2\right] &\leq \frac{1}{n_h^2} \sum_{i \in [n_h]} \mathbb{E}\left[\left\|x^{(i)}_{t+\frac{3}{4}} - \bar{x}_{t+\frac{2}{4}}\right\|^2\right] \\
&\leq \frac{\beta_{hh}}{n_h} \cdot \frac{1}{n_h} \sum_{i \in [n_h]} \left\|x^{(i)}_{t+\frac{2}{4}} - \bar{x}_{t+\frac{2}{4}}\right\|^2
\end{align*}

where the second inequality uses (\ref{eqn: weired-glitch}). Combining this with~\cref{lemma-5} then concludes the proof.
\end{proof}

\paragraph{Stage 3 mixing: Hub-to-Spoke Pull}
Here we have: 
\begin{align*}
    \mathbb{E} \left[ \left\| \bar{x}_{t+1} - \bar{x}_{t+\frac{3}{4}} \right\|^2 \right] 
    =\frac{\beta_{sh}}{n_s n_h}\sum_{i} \mathbb{E} \left[ \left\| x_{t+\frac{3}{4}}^{(i)} - \bar{x}_{t+\frac{3}{4}} \right\|^2 \right]
\end{align*}
\begin{proof}
Note that we can expand the norm as follows:
\begin{align}
    \mathbb{E} \left[ \left\| \bar{y} - \bar{x} \right\|^2 \right] &= \mathbb{E} \left[ \left\| \frac{1}{n} \sum_{i} y^{(i)} - \bar{x} \right\|^2 \right]\nonumber \\ 
    &= \frac{1}{n^2} \sum_{i} \mathbb{E} \left[ \left\| y^{(i)} - \bar{x} \right\|^2 \right] + \frac{1}{n^2} \sum_{i \neq j} \mathbb{E} \left[ \left< y^{(i)} - \bar{x}, y^{(j)} - \bar{x} \right> \right] 
\label{eqn: lemma-1c-begin-copy}
\end{align}
For the third stage of communication from hubs to spokes, we denote ${x}_{t+1}$ as $y$ and ${x}_{t+\frac{3}{4}}$ as $x$, replacing $n$ with $n_s$. For the $i$-th spoke, we have:
\begin{align*}
    x_{t+1}^{(i)} - \bar{x}_{t+\frac{3}{4}} = \frac{1}{b_{sh}} \sum_{k} \mathcal{I}_{k}^{(i)} (x_{t+\frac{3}{4}}^{(k)} - \bar{x}_{t+\frac{3}{4}})
\end{align*}
where $\mathcal{I}_{k}^{(i)}$ is an indicator function that represents the connectivity between spoke $i$ and hub $k$.
We can thus write the second term in (\ref{eqn: lemma-1c-begin-copy}) as

\begin{align}
\frac{1}{n^2}\sum_{i \neq j} \mathbb{E}\left[\langle y^{(i)} - \bar{x}, y^{(j)} - \bar{x} \rangle\right] &= \frac{1}{n_s^2}\sum_{\substack{i \in [n_s] \\i \neq j}} \mathbb{E}\left[\langle x_{t+1}^{(i)} - \bar{x}_{t+\frac{3}{4}}, x_{t+1}^{(j)} - \bar{x}_{t+\frac{3}{4}} \rangle\right] \nonumber \\
&= 
\frac{2}{n_s^2} \sum_{i \ne j} \sum_{k \in [n_h]} \sum_{l \in [n_h]} \mathbb{E} \left[ \frac{\mathcal{I}_{k}^{(i)} \mathcal{I}_{l}^{(j)}}{b_{sh}^2} \langle x_{t+\frac{3}{4}}^{(k)} - \bar{x}_{t+\frac{3}{4}}, x_{t+\frac{3}{4}}^{(l)} - \bar{x}_{t+\frac{3}{4}} \rangle \right]
\label{eqn: lemma-1c-i-kl-copy}
\end{align}

Now note that by symmetry, for any $i, j \in [n_s]$, we have,

\begin{align*}
    \mathbb{E}\left[\mathcal{I}_{k}^{(i)}\mathcal{I}_{l}^{(j)}\right] = \mathbb{E}\left[\mathcal{I}_{3}^{(1)}\mathcal{I}_{4}^{(2)}\right]
\end{align*}

This implies that all three terms in (\ref{eqn: lemma-1c-i-kl-copy}) can be written as,

\begin{align*}
    &c \cdot \mathbb{E} \left[\sum_{k \in [n_h]} \sum_{l \in [n_h]} \left\langle x_{t+\frac{3}{4}}^{(k)} - \bar{x}_{t+\frac{3}{4}}, x_{t+\frac{3}{4}}^{(l)} - \bar{x}_{t+\frac{3}{4}} \right\rangle \right]\\
    &=c \cdot \mathbb{E} \left[ \sum_l \left\| x_{t+\frac{3}{4}}^{(l)} - \bar{x}_{t+\frac{3}{4}} \right\|^2 + \sum_{k \ne l} \left\langle x_{t+\frac{3}{4}}^{(k)} - \bar{x}_{t+\frac{3}{4}}, x_{t+\frac{3}{4}}^{(l)} - \bar{x}_{t+\frac{3}{4}} \right\rangle \right] \\
    &= c \cdot \mathbb{E} \left[ \sum_l \left\| x_{t+\frac{3}{4}}^{(l)} - \bar{x}_{t+\frac{3}{4}} \right\|^2 - \sum_l \left\| x_{t+\frac{3}{4}}^{(l)} - \bar{x}_{t+\frac{3}{4}} \right\|^2\right] \\
    &= 0
\end{align*}

Therefore, from equation (\ref{eqn: lemma-1c-begin-copy}), we obtain
\begin{align}
    \mathbb{E} \left[ \left\| \bar{x}_{t+1} - \bar{x}_{t+\frac{3}{4}} \right\|^2 \right] &= \frac{1}{n_s^2} \sum_{i} \mathbb{E} \left[ \left\| x_{t+1}^{(i)} - \bar{x}_{t+\frac{3}{4}} \right\|^2 \right] \nonumber \\
    & =\frac{\beta_{sh}}{n_s n_h}\sum_{i} \mathbb{E} \left[ \left\| x_{t+\frac{3}{4}}^{(i)} - \bar{x}_{t+\frac{3}{4}} \right\|^2 \right]
\end{align}
where we make use of (\ref{eqn: 1b=iii}).
\end{proof}

\subsection{Proof of~\cref{lemma: 2}}
The expected consensus distance and gradient variance across spokes are bounded as follows:

\begin{enumerate}
    \item \textbf{Consensus Distance Bound:}  
    \begin{align*}
        \frac{1}{n_s^2} \sum_{i,j \in [n_s]} \mathbb{E} \left[ \left\| x_t^{(i)} - x_t^{(j)} \right\|^2 \right] 
        \leq 20 \frac{1 + 3\beta_{HSL}}{(1 - \beta_{HSL})^2} \beta_{HSL} \gamma^2 (\sigma^2 + \mathcal{H}^2).
    \end{align*}

    \item \textbf{Gradient Variance Bound:}  
    \begin{align*}
        \frac{1}{n_s^2} \sum_{i,j \in [n_s]} \mathbb{E} \left[ \left\| g_t^{(i)} - g_t^{(j)} \right\|^2 \right] 
        \leq 15 (\sigma^2 + \mathcal{H}^2).
    \end{align*}
\end{enumerate}
\begin{proof}
For any $i \in [n]$, we have
\begin{align*}
g_{t}^{(i)} - g_{t}^{(j)} &= g_{t}^{(i)} - \nabla f^{(i)}\left(x_{t}^{(i)}\right) + \nabla f^{(i)}\left(x_{t}^{(i)}\right) - \nabla f^{(i)}\left(\bar{x}_t\right) + \nabla f^{(j)}\left(\bar{x}_t\right) \\ &- \nabla f^{(j)}\left(\bar{x}_t\right) + \nabla f^{(j)}\left(\bar{x}_t\right) - \nabla f^{(j)}\left(x_{t}^{(j)}\right) + \nabla f^{(j)}\left(x_{t}^{(j)}\right) - g_{t}^{(j)}
\end{align*}
where $g_{t}$ is the stochastic version of $\nabla f^{(i)}\left(x_{t}^{(i)}\right)$.
Thus, using Jensens's inequality, we have,

\begin{align*}
\left\|g_{t}^{(i)} - g_{t}^{(j)}\right\|^{2} &\leq 5\left\|g_{t}^{(i)} - \nabla f^{(i)}\left(x_{t}^{(i)}\right)\right\|^{2} + 5\left\|\nabla f^{(i)}\left(x_{t}^{(i)}\right) - \nabla f^{(i)}\left(\bar{x}_{t}\right)\right\|^{2} \\
&+ 5\left\|\nabla f^{(j)}\left(x_{t}^{(j)}\right) - \nabla f^{(j)}\left(\bar{x}_{t}\right)\right\|^{2} + 5\left\|g_{t}^{(j)} - \nabla f^{(j)}\left(x_{t}^{(j)}\right)\right\|^{2} \\
&+ 5\left\|\nabla f^{(i)}\left(\bar{x}_{t}\right) - \nabla f^{(j)}\left(\bar{x}_{t}\right)\right\|^{2}
\end{align*}

Taking the conditional expectation, we have,

\begin{align}
\mathbb{E}_{t}\left[\left\|g_{t}^{(i)} - g_{t}^{(j)}\right\|^{2}\right] &\leq 5\mathbb{E}_{t}\left[\left\|g_{t}^{(i)} - \nabla f^{(i)}\left(x_{t}^{(i)}\right)\right\|^{2}\right] + 5\mathbb{E}_{t}\left[\left\|\nabla f^{(i)}\left(x_{t}^{(i)}\right) - \nabla f^{(i)}\left(\bar{x}_{t}\right)\right\|^{2}\right] \nonumber\\
&+ 5\mathbb{E}_{t}\left[\left\|\nabla f^{(j)}\left(x_{t}^{(j)}\right) - \nabla f^{(j)}\left(\bar{x}_{t}\right)\right\|^{2}\right] + 5\mathbb{E}_{t}\left[\left\|g_{t}^{(j)} - \nabla f^{(j)}\left(x_{t}^{(j)}\right)\right\|^{2}\right] \nonumber\\
&+ 5\mathbb{E}_{t}\left[\left\|\nabla f^{(i)}\left(\bar{x}_{t}\right) - \nabla f^{(j)}\left(\bar{x}_{t}\right)\right\|^{2}\right]
\label{eqn:28}
\end{align}

Now by~\cref{ass:bounded_noise}, we have,

\begin{align}
    \mathbb{E}_t \left[ \left\| g_t^{(i)} - \nabla f^{(i)} \left( x_t^{(i)} \right) \right\|^2 \right] \leq \sigma^2
\label{eqn: 29}
\end{align}

By~\cref{ass:smoothness}, we have,

\begin{align}
    \mathbb{E}_{t} \left[ \left\| \nabla f^{(i)}\left(x_{t}^{(i)}\right) - \nabla f^{(i)}\left(\bar{x}_{t}\right) \right\|^2 \right] \leq L^2 \mathbb{E}_{t} \left[ \left\| x_{t}^{(i)} - \bar{x}_{t} \right\|^2 \right]
\label{eqn: 30}
\end{align}

Thus, by~\cref{ass:bounded_heterogeneity}, and~\cref{lemma-5}, we obtain that, 

\begin{align}
    \frac{1}{n_s^2} \sum_{i,j \in [n_s]} \mathbb{E}_t \left[ \left\| \nabla f^{(i)}(\bar{x}_t) - \nabla f^{(j)}(\bar{x}_t) \right\|^2 \right] \leq 2\mathcal{H}^2
\label{eqn: 31}
\end{align}

Combining (\ref{eqn:28}), (\ref{eqn: 29}), (\ref{eqn: 30}), and (\ref{eqn: 31}), and taking total expectation from both sides, we obtain that,

\begin{align}
    \frac{1}{n_s^2} \sum_{i,j \in [n_h]} \mathbb{E} \left[ \left\| g_t^{(i)} - g_t^{(j)} \right\|^2 \right] \leq \frac{10L^2}{n_s} \sum_{i \in [n_s]} \mathbb{E} \left[ \left\| x_t^{(i)} - \bar{x}_t \right\|^2 \right] + 10\sigma^2 + 10\mathcal{H}^2
\label{eqn: 32}
\end{align}

Now~\cref{lemma-5} yields

\begin{align}
    \frac{1}{n_s^2} \sum_{i,j \in [n_s]} \mathbb{E} \left[ \left\| g_t^{(i)} - g_t^{(j)} \right\|^2 \right] \leq \frac{5L^2}{n_s^2} \sum_{i \in [n_s]} \mathbb{E} \left[ \left\| x_t^{(i)} - x_t^{(j)} \right\|^2 \right] + 10\sigma^2 + 10\mathcal{H}^2
\label{eqn: 33}
\end{align}

We now analyze $\frac{1}{n_s^2} \sum_{i,j \in [n_s]} \mathbb{E} \left[ \left\| x_t^{(i)} - x_t^{(j)} \right\|^2 \right]$. From Algorithm 1, recall that for all $i \in [n]$, we have $x_{t+1/4}^{(i)} = x_t^{(i)} - \gamma g_t^{(i)}$. We obtain for all $i, j \in [n_s]$, that

\begin{align}
    \mathbb{E}\left[\left\|x_{t+\frac{1}{4}}^{(i)}-x_{t+\frac{1}{4}}^{(j)}\right\|^2\right] 
    &\leq \mathbb{E} \left[ \left\| x_t^{(i)} - x_t^{(j)} - \gamma (g_t^{(i)} - g_t^{(j)}) \right\|^2 \right] \nonumber \\
    &\leq (1+c)\mathbb{E}\left[\left\|x_t^{(i)}-x_t^{(j)}\right\|^2\right] + \left(1+\frac{1}{c}\right) {\gamma^2}\mathbb{E}\left[\left\|g_t^{(i)}-g_t^{(j)} \right\|^2\right]
\label{eqn: 34}
\end{align}

From (\ref{eqn-1b-final}), we have
\begin{align*}
    \frac{1}{n_h^2}\sum_{i,j} \mathbb{E}\left[\left\|x_{t+\frac{2}{4}}^{(i)} - x_{t+\frac{2}{4}}^{(j)}\right\|^2\right] \leq \frac{\beta_{hs}}{n_s^2} \sum_{i,j} \mathbb{E}\left[\left\|x_{t+\frac{1}{4}}^{(i)} - x_{t+\frac{1}{4}}^{(j)}\right\|^2\right]
\end{align*}

Combining with (\ref{eqn: 34}), we get
\begin{align}
    \frac{1}{n_{h}^2}\sum_{i,j}\mathbb{E}\left[\left\|x_{t+\frac{2}{4}}^{(i)}-x_{t+\frac{2}{4}}^{(j)}\right\|^2\right] &\leq \left(1+c\right)\frac{\beta_{hs}}{n_{s}^2}\sum_{i,j}\mathbb{E}\left[\left\|x_{t}^{(i)}-x_{t}^{(j)}\right\|^2\right] \nonumber \\
&+ \left(1+\frac{1}{c}\right)\gamma^2\frac{\beta_{hs}}{n_{s}^2}\sum_{i,j}\mathbb{E}\left[\left\|g_{t}^{(i)}-g_{t}^{(j)}\right\|^2\right]
\label{eqn: oneplusc}
\end{align}

From (\ref{eqn: weired-glitch}), we also have 
\begin{align*}
    \frac{1}{n_{h}^2}\sum_{i,j}\mathbb{E}\left[\left\|x_{t+\frac{3}{4}}^{(i)}-x_{t+\frac{3}{4}}^{(j)}\right\|^2\right] \leq \frac{\beta_{hh}}{n_{h}^2}\sum_{i,j}\mathbb{E}\left[\left\|x_{t+\frac{2}{4}}^{(i)}-x_{t+\frac{2}{4}}^{(j)}\right\|^2\right]
\end{align*}
We substitute (\ref{eqn: oneplusc}) here to get:
\begin{align}
    \frac{1}{n_{h}^2}\sum_{i,j}\mathbb{E}\left[\left\|x_{t+\frac{3}{4}}^{(i)}-x_{t+\frac{3}{4}}^{(j)}\right\|^2\right] &\leq \left(1+c\right)\frac{\beta_{hs}\beta_{hh}}{n_{s}^2}\sum_{i,j}\mathbb{E}\left[\left\|x_{t}^{(i)}-x_{t}^{(j)}\right\|^2\right] \\
&+ \left(1+\frac{1}{c}\right)\gamma^2\frac{\beta_{hs} \beta_{hh}}{n_{s}^2}\sum_{i,j}\mathbb{E}\left[\left\|g_{t}^{(i)}-g_{t}^{(j)}\right\|^2\right]
\label{eqn: lost-count}
\end{align}
Also, from (\ref{eqn: 1b-iii-in-2}), we obtain
\begin{align*}
    \frac{1}{n_{s}^2}\sum_{i,j}\mathbb{E}\left[\left\|x_{t+1}^{(i)}-x_{t+1}^{(j)}\right\|^2\right] \leq \frac{\beta_{hs}}{n_{s}^2}\sum_{i,j}\mathbb{E}\left[\left\|x_{t+\frac{3}{4}}^{(i)}-x_{t+\frac{3}{4}}^{(j)}\right\|^2\right]
\end{align*}
We combine this with (\ref{eqn: lost-count}) to get
\begin{align*}
    \frac{1}{n_{s}^2}\sum_{i,j}\mathbb{E}\left[\left\|x_{t+1}^{(i)}-x_{t+1}\right\|^2\right] &\leq \left(1+c\right)\frac{\beta_{HSL}}{n_{s}^2}\sum_{i,j}\mathbb{E}\left[\left\|x_{t}^{(i)}-x_{t}^{(j)}\right\|^2\right] \\
    &+ \left(1+\frac{1}{c}\right)\gamma^2\frac{\beta_{HSL}}{n_{s}^2}\sum_{i,j}\mathbb{E}\left[\left\| g_{t}^{(i)}-g_{t}^{(j)}\right\|^2\right]
\end{align*}

For  $c = \frac{1 - \beta_{HSL}}{4\beta_{HSL}}$, we obtain that,

\begin{align*}
    \frac{1}{n_s^2} \sum_{i,j \in [n_s]} \mathbb{E} \left[ \left\| x_{t+1}^{(i)} - x_{t+1}^{(j)} \right\|^2 \right] &\le \frac{1 + 3\beta_{HSL}}{4} \frac{1}{n_s^2} \sum_{i,j \in [n_s]} \mathbb{E} \left[ \left\| x_t^{(i)} - x_t^{(j)} \right\|^2 \right]\\
    &+ \frac{1 + 3\beta_{HSL}}{1 - \beta_{HSL}} \beta_{HSL} \gamma^2 \frac{1}{n_s^2} \sum_{i,j \in [n_s]} \mathbb{E} \left[ \left\| g_t^{(i)} - g_t^{(j)} \right\|^2 \right]
\end{align*}

Combining this with (\ref{eqn: 33}), we obtain that

\begin{align*}
    \frac{1}{n_s^2} \sum_{i,j \in [n_s]} \mathbb{E} \left[ \left\| x_{t+1}^{(i)} - x_{t+1}^{(j)} \right\|^2 \right] &\le \frac{1 + 3\beta_{HSL}}{4} \frac{1}{n_s^2} \sum_{i,j \in [n_s]} \mathbb{E} \left[ \left\| x_t^{(i)} - x_t^{(j)} \right\|^2 \right]\\
    &+ \frac{1 + 3\beta_{HSL}}{1 - \beta_{HSL}} \beta_{HSL} \gamma^2 \left( \frac{5L^2}{n_s^2} \sum_{i,j \in [n_s]} \mathbb{E} \left[ \left\| x_t^{(i)} - x_t^{(j)} \right\|^2 \right] + 10\sigma^2 + 10\mathcal{H}^2 \right)\\
    &= \left( \frac{1 + 3\beta_{HSL}}{4} + 5\frac{1 + 3\beta_{HSL}}{1 - \beta_{HSL}}\beta_{HSL}\gamma^2 L^2 \right) \frac{1}{n_s^2} \sum_{i,j \in [n_s]} \mathbb{E} \left[ \left\| x_t^{(i)} - x_t^{(j)} \right\|^2 \right]\\
    &+ \frac{1 + 3\beta_{HSL}}{1 - \beta_{HSL}} \beta_{HSL} \gamma^2 (10\sigma^2 + 10\mathcal{H}^2).
\end{align*}

Now note that from~\cref{remark: 2} we have $\beta_{HSL} \leq 1 - \frac{1}{e}$ which implies that

\begin{align*}
    \gamma^2 \le \frac{1}{(20L)^2} \le \frac{(1 - \beta_{HSL})^2}{20\beta_{HSL}(1 + 3\beta_{HSL})L^2}
\end{align*}

Therefore,

\begin{align*}
    \frac{1}{n_s^2} \sum_{i,j \in [n_s]} \mathbb{E} \left[ \left\| x_{t+1}^{(i)} - x_{t+1}^{(j)} \right\|^2 \right] \le \frac{1 + \beta_{HSL}}{2} \frac{1}{n_s^2} \sum_{i,j \in [n_s]} \mathbb{E} \left[ \left\| x_t^{(i)} - x_t^{(j)} \right\|^2 \right] + \frac{1 + 3\beta_{HSL}}{1 - \beta_{HSL}} \beta_{HSL} \gamma^2 (10\sigma^2 + 10\mathcal{H}^2).
\end{align*}

Unrolling the recursion, we obtain that,

\begin{align*}
    \frac{1}{n_s^2} \sum_{i,j \in [n_s]} \mathbb{E} \left[ \left\| x_t^{(i)} - x_t^{(j)} \right\|^2 \right] \le 20 \frac{1 + 3\beta_{HSL}}{(1 - \beta_{HSL})^2} \beta_{HSL} \gamma^2 (\sigma^2 + \mathcal{H}^2).
\end{align*}

Combining this with (\ref{eqn: 32}), we obtain that,

\begin{align*}
    \frac{1}{n_s^2} \sum_{i,j \in [n_s]} \mathbb{E} \left[ \left\| g_t^{(i)} - g_t^{(j)} \right\|^2 \right] \le 15 (\sigma^2 + \mathcal{H}^2).
\end{align*}

\end{proof}

\subsubsection{Proof of~\cref{lemma: 3}}
The expected gradient norm of the global objective satisfies the following upper bound:
\begin{align*}
    \mathbb{E}\left[\left\|\nabla F(\bar{x}_t)\right\|^2\right] 
    &\leq \frac{2}{\gamma}\mathbb{E}\left[F(\bar{x}_t) - F(\bar{x}_{t+1})\right] 
    + \frac{L}{2 n^2}\sum_{i,j} \mathbb{E}\left[\left\|x_t^{(i)} - x_t^{(j)}\right\|^2\right] 
    + \frac{4L\gamma \sigma^2}{n} \\
    &\quad + \frac{4L}{\gamma} \mathbb{E} \Big[ \left\|\bar{x}_{t+1} - \bar{x}_{t+\frac{3}{4}}\right\|^2 
    + \left\|\bar{x}_{t+\frac{3}{4}} - \bar{x}_{t+\frac{2}{4}}\right\|^2 
    + \left\|\bar{x}_{t+\frac{2}{4}} - \bar{x}_{t+\frac{1}{4}}\right\|^2 \Big].
\end{align*}
\begin{proof}
Consider an arbitrary $t \in [T]$. Then from the smoothness property, we have:
\begin{align}
F(\bar{x}_{t+1}) - F(\bar{x}_t) &\le \langle \bar{x}_{t+1} - \bar{x}_t, \nabla F(\bar{x}_t) \rangle + \frac{L}{2} ||\bar{x}_{t+1} - \bar{x}_t||^2 \nonumber \\
&= \langle \bar{x}_{t+1} - \bar{x}_{t+\frac{3}{4}} + \bar{x}_{t+\frac{3}{4}}- \bar{x}_{t+\frac{2}{4}} + \bar{x}_{t+\frac{2}{4}} - \bar{x}_{t+\frac{1}{4}} + \bar{x}_{t+\frac{1}{4}} - \bar{x}_t, \nabla F(\bar{x}_t) \rangle \nonumber \\
&+ \frac{L}{2} ||\bar{x}_{t+1} - \bar{x}_{t+\frac{3}{4}} + \bar{x}_{t+\frac{3}{4}}- \bar{x}_{t+\frac{2}{4}} + \bar{x}_{t+\frac{2}{4}} - \bar{x}_{t+\frac{1}{4}} + \bar{x}_{t+\frac{1}{4}} - \bar{x}_t||^2.
\label{eqn: 3-0}
\end{align}

From~\cref{lemma: 1a}, we have
\[
\mathbb{E}\big[\overline{x}_{t+1}\big] = \mathbb{E}\big[\overline{x}_{t+\frac{3}{4}}\big] = \mathbb{E}\big[\overline{x}_{t+\frac{2}{4}}\big] = \mathbb{E}\big[\overline{x}_{t+\frac{1}{4}}\big].
\]
Now, we take conditional expectation on (\ref{eqn: 3-0}) and use~\cref{lemma: 1a} to get
\begin{align}
\mathbb{E}_t \left[F(\bar{x}_{t+1}) - F(\bar{x}_t)\right] 
&\leq \langle \mathbb{E}_t \left[\bar{x}_{t+1} - \bar{x}_t\right], \nabla F(\bar{x}_t) \rangle 
+ \frac{L}{2} \mathbb{E}_t \left[||\bar{x}_{t+1} - \bar{x}_{t+\frac{3}{4}} + \bar{x}_{t+\frac{3}{4}}- \bar{x}_{t+\frac{2}{4}} + \bar{x}_{t+\frac{2}{4}} - \bar{x}_{t+\frac{1}{4}} + \bar{x}_{t+\frac{1}{4}} - \bar{x}_t||^2\right] \nonumber \\
&\leq -\gamma \langle \overline{\nabla F}_t, \nabla F(\bar{x}_t)\rangle + 2L\gamma^2 \mathbb{E}_t \left[\left\|\bar{g}_t \right\|^2 \right] 
+ 2L \, \mathbb{E}_t \left[ \left\| \bar{x}_{t+1} - \bar{x}_{t+\frac{3}{4}} \right\|^2 \right] + 2L \, \mathbb{E}_t \left[ \left\| \bar{x}_{t+\frac{3}{4}} - \bar{x}_{t+\frac{2}{4}} \right\|^2 \right] \nonumber \\
&+ 2L \, \mathbb{E}_t \left[ \left\| \bar{x}_{t+\frac{2}{4}} - \bar{x}_{t+\frac{1}{4}} \right\|^2 \right]
\label{eqn: 3-1}
\end{align}
where 
$
\overline{\nabla F}_t = \frac{1}{n_s} \sum_{i \in [n_s]} \nabla f^{(i)} (x_t^{(i)}), \quad \bar{g}_t = \frac{1}{n_s} \sum_{i \in [n_s]} g_t^{(i)}
$, and we make use of Jensen's inequality.
Then we use the $\mathbb{E}_t[\bar{g}_t] = \overline{\nabla F}_t$, and $\mathbb{E}_t[\left\|\bar{g}_t - \overline{\nabla F}_t\right\|^2] \leq \frac{\sigma^2}{n}$

\begin{align}
    \mathbb{E}_t \left[F(\bar{x}_{t+1}) - F(\bar{x}_t)\right] &\leq -\gamma \langle \overline{\nabla F}_t, \nabla F(\bar{x}_t)\rangle + 2L\gamma^2 \mathbb{E} \left[ \left\| \nabla F(\bar{x}_t) \right\|^2 \right] + 2L \frac{\gamma^2 \sigma^2}{n} 
+ 2L \, \mathbb{E}_t \left[ \left\| \bar{x}_{t+1} - \bar{x}_{t+\frac{3}{4}} \right\|^2 \right] \nonumber \\
&+ 2L \, \mathbb{E}_t \left[ \left\| \bar{x}_{t+\frac{3}{4}} - \bar{x}_{t+\frac{2}{4}} \right\|^2 \right] 
+ 2L \, \mathbb{E}_t \left[ \left\| \bar{x}_{t+\frac{2}{4}} - \bar{x}_{t+\frac{1}{4}} \right\|^2 \right] 
\label{eqn: 35}
\end{align}
Then we use $\gamma \leq \frac{1}{4L}$ to get
\begin{align}
-\gamma\langle\overline{\nabla F}_{t},\nabla F(\overline{x}_{t})\rangle+2L\gamma^{2}||\overline{\nabla F}_{t}||^{2} &\le\frac{\gamma}{2}(-2\langle\overline{\nabla F}_{t},\nabla F(\overline{x}_{t})\rangle+||\overline{\nabla F}_{t}||^{2}) \nonumber \\
&= \frac{\gamma}{2}(-||\nabla F(\overline{x}_{t})||^{2}+||\nabla F(\overline{x}_{t})-\overline{\nabla F}_{t}||^{2}).
\label{eqn: 36}
\end{align}
Combining (\ref{eqn: 35}) and (\ref{eqn: 36}), we obtain

\begin{align*}
    \mathbb{E}_t \left[ F\left(\bar{x}_{t+1}\right) - F\left(\bar{x}_t\right) \right] &\leq -\frac{\gamma}{2}  \left\| \nabla F\left(\bar{x}_t\right) \right\|^2 + \frac{\gamma}{2} \left\| \nabla F(\bar{x}_t) - \bar{\nabla F_t} \right\|^2 + 2L \frac{\gamma^2 \sigma^2}{n}\\
    &+ 2L \mathbb{E}_t \left[ \left\| \bar{x}_{t+1} - \bar{x}_{t+\frac{3}{4}} \right\|^2 \right] 
    + 2L \mathbb{E}_t \left[ \left\| \bar{x}_{t+\frac{3}{4}} - \bar{x}_{t+\frac{2}{4}} \right\|^2 \right] 
    + 2L \mathbb{E}_t \left[ \left\| \bar{x}_{t+\frac{2}{4}} - \bar{x}_{t+\frac{1}{4}} \right\|^2 \right]
\end{align*}
Taking total expectation, we obtain
\begin{align}
    \mathbb{E}_t \left[ F\left(\bar{x}_{t+1}\right) - F\left(\bar{x}_t\right) \right] &\leq -\frac{\gamma}{2} \mathbb{E}_t \left[ \left\| \nabla F\left(\bar{x}_t\right) \right\|^2 \right] + \frac{\gamma}{2} \mathbb{E}_t \left[ \left\| \nabla F(\bar{x}_t) - \bar{\nabla F_t} \right\|^2 \right] + 2L \frac{\gamma^2 \sigma^2}{n} \nonumber \\
    &+ 2L \mathbb{E}_t \left[ \left\| \bar{x}_{t+1} - \bar{x}_{t+\frac{3}{4}} \right\|^2 \right] 
    + 2L \mathbb{E}_t \left[ \left\| \bar{x}_{t+\frac{3}{4}} - \bar{x}_{t+\frac{2}{4}} \right\|^2 \right] 
    + 2L \mathbb{E}_t \left[ \left\| \bar{x}_{t+\frac{2}{4}} - \bar{x}_{t+\frac{1}{4}} \right\|^2 \right]
\label{eqn: 37}
\end{align}
Now, note that

\begin{align*}
    \mathbb{E} \left[ \left\| \overline{\nabla F}_t - \nabla F(\overline{x}_t) \right\|^2 \right] &= \mathbb{E} \left[ \left\| \frac{1}{n_s} \sum_{i \in [n_s]} \nabla f^{(i)}(x_t^{(i)}) - \frac{1}{n_s} \sum_{i \in [n_s]} \nabla f^{(i)}(\overline{x}_t) \right\|^2 \right]\\
    &= \mathbb{E} \left[ \left\| \frac{1}{n_s} \sum_{i \in [n_s]} \left( \nabla f^{(i)}(x_t^{(i)}) - \nabla f^{(i)}(\overline{x}_t) \right) \right\|^2 \right]\\
    &\le \frac{1}{n_s} \sum_{i \in [n_s]} \mathbb{E} \left[ \left\| \nabla f^{(i)}(x_t^{(i)}) - \nabla f^{(i)}(\overline{x}_t) \right\|^2 \right] \\
&\le \frac{L^2}{n_s} \sum_{i \in [n_s]} \mathbb{E} \left[ \left\| x_t^{(i)} - \overline{x}_t \right\|^2 \right]. \\
&\le \frac{L^2}{2n_s^2} \sum_{i,j \in [n_s]} \mathbb{E} \left[ \left\| x_t^{(i)} - x_t^{(j)} \right\|^2 \right],
\end{align*}
where we use~\cref{ass:smoothness} and~\cref{lemma-5} in the above two inequalities. Now, substituting this back in (\ref{eqn: 37}).
\begin{align*}
    \mathbb{E}_t \left[ F\left(\bar{x}_{t+1}\right) - F\left(\bar{x}_t\right) \right] &\leq -\frac{\gamma}{2} \mathbb{E}_t \left[ \left\| \nabla F\left(\bar{x}_t\right) \right\|^2 \right] + \frac{\gamma}{2} \frac{L^2}{2n_s^2} \sum_{i,j \in [n_s]} \mathbb{E} \left[ \left\| x_t^{(i)} - x_t^{(j)} \right\|^2 \right] + 2L \frac{\gamma^2 \sigma^2}{n}\\
    &+ 2L \mathbb{E}_t \left[ \left\| \bar{x}_{t+1} - \bar{x}_{t+\frac{3}{4}} \right\|^2 \right] 
    + 2L \mathbb{E}_t \left[ \left\| \bar{x}_{t+\frac{3}{4}} - \bar{x}_{t+\frac{2}{4}} \right\|^2 \right] 
    + 2L \mathbb{E}_t \left[ \left\| \bar{x}_{t+\frac{2}{4}} - \bar{x}_{t+\frac{1}{4}} \right\|^2 \right]
\end{align*}
Rearranging these terms, we get
\begin{align*}
    \mathbb{E}\left[\left\|\nabla F(\bar{x}_t)\right\|^2\right] \leq \frac{2}{\gamma}\mathbb{E}\left[F(\bar{x}_t) - F(\bar{x}_{t+1})\right] + \frac{L}{2 n^2}\sum_{i,j} \mathbb{E}\left[\left\|x_t^{(i)} - x_t^{(j)}\right\|^2\right] + \frac{4L\gamma \sigma^2}{n} \\
    + \frac{4L}{\gamma}\mathbb{E}\left[\left\|\bar{x}_{t+1} - \bar{x}_{t+\frac{3}{4}}\right\|^2 + \left\|\bar{x}_{t+\frac{3}{4}} - \bar{x}_{t+\frac{2}{4}}\right\|^2 + \left\|\bar{x}_{t+\frac{2}{4}} - \bar{x}_{t+\frac{1}{4}}\right\|^2\right]
\end{align*}

\end{proof}
\subsection{Proof of~\cref{lemma-5}} 

For any set $\{x_t^{(i)}\}_{i \in [n]}$ of $n$ vectors, we have
\[
\frac{1}{n} \sum_{i \in [n]} \|x_t^{(i)} - \bar{x}_t\|^2 = \frac{1}{2n^2} \sum_{i, j \in [n]} \|x_t^{(i)} - x_t^{(j)}\|^2,
\]
where $\bar{x}_t = \frac{1}{n} \sum_{i \in [n]} x_t^{(i)}$.

\textbf{Proof}
\begin{align*}
\frac{1}{n^2} \sum_{i, j \in [n]} \|x_t^{(i)} - x_t^{(j)}\|^2 &= \frac{1}{n^2} \sum_{i, j \in [n]} \|(x_t^{(i)} - \bar{x}_t) - (x_t^{(j)} - \bar{x}_t)\|^2 \\
&= \frac{1}{n^2} \sum_{i, j \in [n]} \left[ \|x_t^{(i)} - \bar{x}_t\|^2 + \|x_t^{(j)} - \bar{x}_t\|^2 - 2 \langle x_t^{(i)} - \bar{x}_t, x_t^{(j)} - \bar{x}_t \rangle \right] \\
&= \frac{2}{n} \sum_{i \in [n]} \|x_t^{(i)} - \bar{x}_t\|^2 - \frac{2}{n^2} \sum_{i \in [n]} \left\langle x_t^{(i)} - \bar{x}_t, \sum_{j \in [n]} (x_t^{(j)} - \bar{x}_t) \right\rangle.
\end{align*}
Noting that $\sum_{j \in [n]} (x_t^{(j)} - \bar{x}_t) = 0$, yields the desired result.

\end{document}